\documentclass[10pt,journal,compsoc]{IEEEtran}

\usepackage[nocompress]{cite}

\usepackage{url}
\usepackage{graphicx}
\usepackage{subcaption}
\usepackage{enumitem}
\usepackage{amsmath}
\usepackage{amssymb}

\usepackage{booktabs}
\usepackage{colortbl} 
\usepackage{xcolor}
\usepackage{bigstrut}
\usepackage{multirow}
\usepackage{rotating} 
\usepackage{diagbox}
\usepackage{color}

\usepackage{algorithm}  
\usepackage{algorithmicx}  
\usepackage{algpseudocode}  
\floatname{algorithm}{Algorithm}

\newcommand{\etal}{\textit{et al. }}
\newcommand{\ie}{\textit{i.e.}}

\newcommand{\etc}{\textit{etc}}

\begin{document}
\title{Pixel Distillation: Cost-flexible Distillation across Image Sizes and Heterogeneous Networks}

\author{Guangyu Guo,
	Dingwen Zhang,~\IEEEmembership{Member,~IEEE}
	Longfei Han,
	Nian Liu,
	Ming-Ming Cheng,~\IEEEmembership{Senior Member,~IEEE}
	and Junwei Han,~\IEEEmembership{Fellow,~IEEE}
	\IEEEcompsocitemizethanks{
        \IEEEcompsocthanksitem This work is partially supported by National Science and Technology Major Project (2022ZD0119004), National Natural
Science Foundation of China (No.62202015, 62293543, 62322605, U21B2048), Anhui Provincial Key R\&D Programmes (2023s07020001), the University Synergy Innovation Program of Anhui Province (GXXT-2022-052).
		\IEEEcompsocthanksitem Guangyu Guo is with Brain and Artificial Intelligence Laboratory, School of Automation, Northwestern Polytechnical University, Xi'an, China. Email: gyguo95@gmail.com
		\IEEEcompsocthanksitem Dingwen Zhang is with the Brain and Artificial Intelligence Laboratory, School of Automation, Northwestern Polytechnical University, Xi'an, China, and also with Xijing Hospital, The Fourth Military Medical University, Xi'an, China. Email: zhangdingwen2006yyy@gmail.com
		\IEEEcompsocthanksitem Longfei Han is with Beijing Technology And Business University, Beijing, China. Email: longfeihan@btbu.edu.cn
		\IEEEcompsocthanksitem Nian Liu is with the Mohamed bin Zayed University of Artificial Intelligence, Abu Dhabi, United Arab. Emirates. E-mail: liunian228@gmail.com
		\IEEEcompsocthanksitem Ming-Ming Cheng is with TKLNDST, College of Computer Science, Nankai University, Tianjin, China. Email:cmm@nankai.edu.cn
		\IEEEcompsocthanksitem Junwei Han is with the Hefei Comprehensive National Science Center, Institute of Artificial Intelligence, Hefei, China. Email: junweihan2010@gmail.com
		\IEEEcompsocthanksitem Corresponding author: Dingwen Zhang, Junwei Han.}
}

\IEEEtitleabstractindextext{%
\begin{abstract}
Previous knowledge distillation (KD) methods mostly focus on compressing network architectures, which is not thorough enough in deployment as some costs like transmission bandwidth and imaging equipment are related to the image size. Therefore, we propose Pixel Distillation that extends knowledge distillation into the input level while simultaneously breaking architecture constraints. Such a scheme can achieve flexible cost control for deployment, as it allows the system to adjust both network architecture and image quality according to the overall requirement of resources. Specifically, we first propose an input spatial representation distillation (ISRD) mechanism to transfer spatial knowledge from large images to student's input module, which can facilitate stable knowledge transfer between CNN and ViT. Then, a Teacher-Assistant-Student (TAS) framework is further established to disentangle pixel distillation into the model compression stage and input compression stage, which significantly reduces the overall complexity of pixel distillation and the difficulty of distilling intermediate knowledge. Finally, we adapt pixel distillation to object detection via an aligned feature for preservation (AFP) strategy for TAS, which aligns output dimensions of detectors at each stage by manipulating features and anchors of the assistant.
Comprehensive experiments on image classification and object detection demonstrate the effectiveness of our method. Code is available at \url{https://github.com/gyguo/PixelDistillation}.
\end{abstract}

}

\maketitle
\IEEEdisplaynontitleabstractindextext
\IEEEpeerreviewmaketitle

\IEEEraisesectionheading{\section{Introduction}\label{sec:introduction}}

\IEEEPARstart{R}{ecently}, great success has been made in the computer vision community based on the rapid development of CNNs~\cite{simonyan2015very,he2016deep,zhang2018shufflenet,guo2022visual,Liu_2022_CVPR}, ViTs~\cite{dosovitskiy2020vit,liu2021swin,touvron2021training} and foundation models~\cite{kirillov2023segment,wang2023seggpt,zou2023segment}. While these models have been able to achieve very promising performance on high-performance computing devices, it is hard to equip them on edge devices like smartphones, embedded devices, small-size UAVs, \etc. This is because these approaches are usually designed with complex network architectures and large-scale network parameters, while some edge devices require lower transmission bandwidth and computing resources. 

\begin{figure}[!t]
	\begin{center}
		\includegraphics[width=1\linewidth]{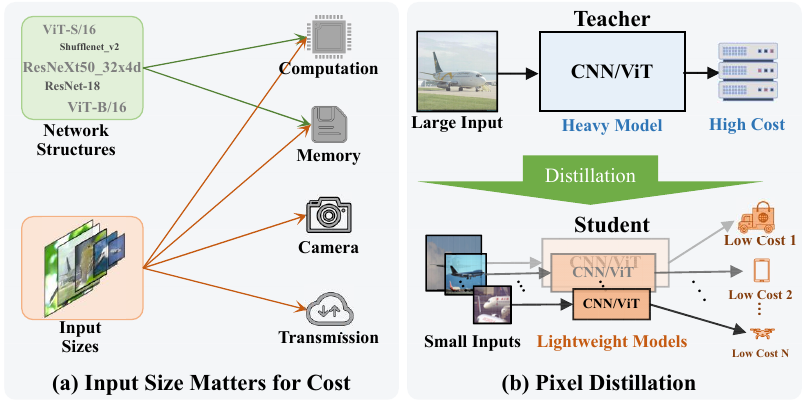}
	\end{center}
	\caption{
		(a) Compared to network architecture, input size has an impact on more kinds of costs, including requirements for cameras and transmission bandwidth. (b) Pixel distillation can provide more flexible cost control schemes for deployment by distilling knowledge across different input sizes and heterogeneous networks.}
	\label{fig_intro}
\end{figure}

\begin{figure*}[!t]
	\centering
	\includegraphics[width=1\linewidth]{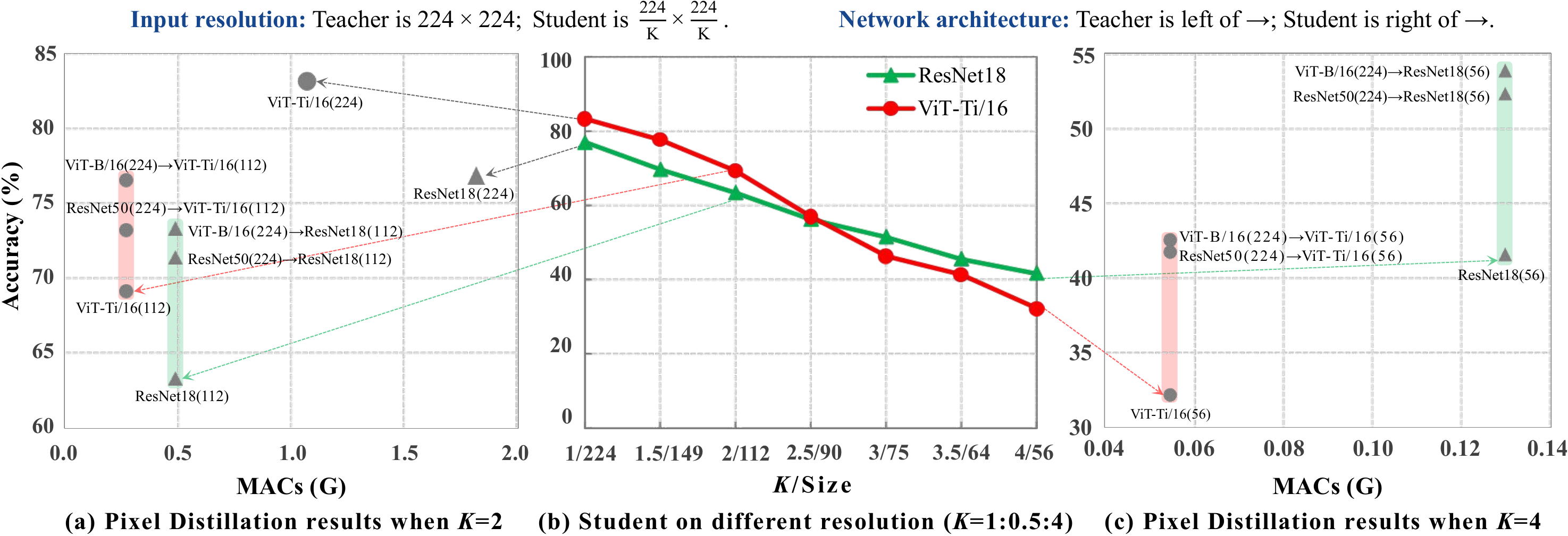}
	\caption{Observations about pixel distillation. We report the accuracy (\%) of baseline ResNet18 and ViT-Ti/16 under different input sizes in figure b, and the accuracy (\%) \textit{vs.} MACs (G) of our pixel distillation method when the input size is $112\times 112$ (figure a) and $56\times 56$ (figure c), respectively. \textbf{(b)} reports the performance of two student networks under seven input resolutions. \textbf{(a)} and \textbf{{(c)}} report the performance of our pixel distillation method under two input resolution settings, \ie, $K=2$ and $K=4$. The arrows drawn from (b) represent the baseline performance without knowledge distillation.
 } 
\label{fig:show}
\end{figure*}

To deal with this situation, KD techniques that aim at using smaller network architectures received great attention in the past few years---usually with fewer network layers or smaller channel dimensions, thus reducing the requirement in computation. However, besides the internal network architecture, the external factor, \ie, the input size, reminds us that the existing research is not sufficient. As we illustrated in Fig.~\ref{fig_intro}a, besides the computational complexity and running memory, the input size also matters for the costs of transmission and imaging equipment. For example, if the width and height of an image become $K$ times smaller, the network would only require approximately $1/K^{2}$ of the original computational complexity and running memory. Moreover, for the cases where the computation is completed on the remote server, the transmission cost will also be greatly reduced by using a small input size. Meanwhile, in many real-world applications like some embedded systems, the devices might be only equipped with low-resolution cameras to keep a low cost of equipment. In conclusion, there is an urgent demand to enable well-trained deep models to fit on small images.  

To solve these problems, we propose a new distillation framework, called pixel distillation, to achieve the best trade-off between the performance and the cost including the computational complexity, storage, and transmission. As shown in Fig.~\ref{fig_intro}b, Pixel distillation generalizes the idea of knowledge distillation to the input level, where a large input is used by a heavy teacher model and a small size is used by a lightweight student model. A pixel distillation method should satisfy two criteria. The first is the \textbf{input adaptability criterion:} \textit{the method can adapt to different small sizes for the input of students to guarantee flexibility of cost control schemes}. The second is the \textbf{architecture adaptability criterion:} \textit{the method should be available to various architectures of the teacher and student to obtain better performance, including the case where the teacher and student belong to CNNs and ViTs, respectively.} The necessity of the architecture criterion comes from the phenomenon that different networks have varying adaptability to changes in the input size. As shown in Fig.~\ref{fig:show}b, ViT-Ti/16~\cite{dosovitskiy2020vit} performs better when the input size is large, but its performance decreases faster than ResNet18~\cite{he2016deep} when the input size becomes smaller. When the small size is set as $112 \times 112$, student ViT-Ti/16 can obtain better performance than ResNet18 with less computational complexity (Fig.~\ref{fig:show}a). However, when the input size is reduced to $56 \times 56$, student ResNet18 outperforms ViT-Ti/16 by 10\% in terms of accuracy (Fig.~\ref{fig:show}c). Hence, a pixel distillation method should be adaptive to various network architectures to obtain students with better performance.

In this paper, we first propose a baseline for the proposed pixel distillation scheme for the image classification task, called vanilla PD, which satisfies the aforementioned two criteria (see Fig.~\ref{fig_ird}). To be specific, on the assumption that the small input images in pixel distillation would lead to inadequate spatial information on the shallow features of the student, we propose an Input Spatial Representation Distillation (ISRD) mechanism to distill knowledge from the large input to help the input module of the student obtain richer representation. As illustrated in Fig.~\ref{fig_ird}b, considering that our student network includes not only CNNs but also ViTs, we design a Generalized Spatial Feature Preprocess (GSFP) module as the encoder to transfer input spatial features from CNNs and ViTs into the same form. Moreover, as illustrated in Fig.~\ref{fig_ird}c, the decoder of the ISRD can convert the encoded features of arbitrary volume into a pseudo large image, which makes the ISRD mechanism able to be used under any input size of the student. By combining ISRD with the previous prediction distillation methods, we can obtain a simple one-stage trained baseline method for pixel distillation, \ie, the vanilla PD, which satisfies two necessary criteria. 

Although the vanilla PD is able to achieve sufficiently good results, we hope that feature distillation can be used to further enhance the performance. Moreover, the gap between the teacher and student in pixel distillation not only comes from different model architectures but also varying input resolutions, which is larger than that in traditional knowledge distillation, distilling useful knowledge for the student will be more difficult~\cite{cho2019efficacy,qian2022switchable}. Therefore, we propose a two-stage distillation strategy where an assistant network is introduced into the classical Teacher-Student (TS) framework. As shown in Fig.~\ref{fig_tas}, the proposed Teacher-Assistant-Student framework separates the pixel distillation process into the model compression stage and input compression stage. At each stage, it will be easier for the student to mimic the teacher than that in the one-stage distillation mechanism. Moreover, since the teacher and student have the same architecture in the input compression stage, TAS makes it easier to design a feature distillation mechanism to relieve the performance degradation caused by the small input. 
Finally, when we adapt the concept of pixel distillation to object detection, we observe that the variability in image resolution among object detectors correspondingly affects their output dimensions like the number of anchor boxes, complicating the preservation of knowledge from the teacher's detection head. To address this, we propose a strategy termed Aligned Feature for Preservation (AFP) for the assistant network. This strategy involves integrating an upsampling operation to match the feature dimension of the assistant network with those of the teacher during the first knowledge distillation stage. Subsequently, during knowledge transfer from assistant to student, we remove the upsampling step since both networks handle inputs of the same resolution.
TAS can make the student network effectively leverage knowledge from both the heavy model and large input, and make it flexible to be utilized for more complex tasks.

One of the goals of our research is to evaluate the performance of our proposed method in realistic settings. Therefore, we choose three widely used datasets that reflect different challenges and characteristics of image classification tasks. The first two datasets are CUB-200-2011~\cite{wah2011caltech} and FGCV-aircraft~\cite{maji2013fine}, which contain fine-grained categories of birds and aircraft respectively. The third dataset is ImageNet~\cite{deng2009imagenet}, which is a large-scale dataset with 1000 classes and millions of images. We conduct extensive experiments on these benchmarks to demonstrate the effectiveness and robustness of our method.
Also, to evaluate the pixel distillation paradigm in a more complex task, \ie, object detection, experiments are conducted on Pascal VOC~\cite{everingham2010pascal} and MS-COCO 2017 dataset~\cite{lin2014microsoft}.
As shown in Fig.~\ref{fig:show}, on the CUB-200-2011 dataset, our proposed method can make ViT-Ti/16 with $112\times 112$ input achieve comparable performance with ResNet18 that uses input images of size $224\times 224$ (76.74\% vs. 76.89\%), while only 13\% computational complexity is required (0.273G MACs vs. 1.82G MACs), and only 25\% of the storage and transmission costs are needed.

To summarize, the contribution of this paper is fourfold:
\begin{itemize}
	\item We present a new distillation scheme called pixel distillation, which provides an early attempt to establish a flexible KD scheme for edge devices with small input sizes.
	\item We present an input spatial representation distillation mechanism that adapts to input images of different small sizes and can be applied to common network architectures such as CNNs and ViTs.
	\item We propose a Teacher-Assistant-Student distillation framework that reduces the learning difficulty of the student in pixel distillation and enables feature distillation when the input size of the student is reduced.
 \item We adapt pixel distillation to object detection, crafting an aligned feature preservation strategy for the assistant network to tackle the challenge of inconsistent output dimensions due to varied image resolutions.
	\item Extensive experiments on image classification and object detection demonstrate the effectiveness and efficiency of the proposed distillation scheme.
\end{itemize}

\section{Related Works and Relevance}
In this section, we review several categories of the existing commonly used methods that can transfer knowledge from a strong source model to a weak target model. Besides, in Table~\ref{tab:tasks} we provide the difference between our pixel distillation paradigm and with the previous method from aspects of testing efficiency, performance gain for the student, and adaptability to the input size and network architecture.

\begin{table}[!t]
	\centering
	\resizebox{\linewidth}{!}{
		\begin{tabular}{c|ccccc}
			\hline
			\rowcolor[rgb]{ .922,  .945,  .871}       & FT    & PKD   & FKD   & LR-KD & PD \\
			\hline
			\rowcolor[rgb]{ .773,  .851,  .945} Testing efficiency & fast  & \textcolor[rgb]{ 1,  0,  0}{$\times$slow} & \textcolor[rgb]{ 1,  0,  0}{$\times$slow} & fast  & \textcolor[rgb]{ 0,  .69,  .314}{$\surd$fast} \\
			\rowcolor[rgb]{ .773,  .851,  .945} Performance gain & \textcolor[rgb]{ 1,  0,  0}{$\times$low} & high  & high  & high  & \textcolor[rgb]{ 0,  .69,  .314}{$\surd$high} \\
			\rowcolor[rgb]{ .773,  .851,  .945} Input adaptability & high  & medium  & \textcolor[rgb]{ 1,  0,  0}{$\times$low} & \textcolor[rgb]{ 1,  0,  0}{$\times$low} & \textcolor[rgb]{ 0,  .69,  .314}{$\surd$high} \\
			\rowcolor[rgb]{ .773,  .851,  .945} Architecture adaptability & \textcolor[rgb]{ 1,  0,  0}{$\times$low} & high  & medium & \textcolor[rgb]{ 1,  0,  0}{$\times$low} & \textcolor[rgb]{ 0,  .69,  .314}{$\surd$high} \\
			\hline
		\end{tabular}%
	}
	\caption{Illustration of the difference between the following tasks: fine-tuning (FT), knowledge distillation including prediction-based knowledge distillation (PKD) and feature-based knowledge distillation (FKD), low-resolution recognition with knowledge distillation (LR-KD), our proposed pixel distillation (PD).}
	\label{tab:tasks}%
\end{table}%

\subsection{Knowledge Distillation}
\textbf{Knowledge distillation in image classification:} Hinton \etal first introduced the notion of knowledge distillation which aims to train a smaller model (\ie, student) via learning from the cumbersome models (\ie, teacher)~\cite{hinton2015distilling}. The early knowledge distillation methods used the predicted score of the teacher model to guide the training of the student model. An essential way is to regard the predicted logits of the teacher model as the soft target of the student~\cite{hinton2015distilling,ba2014deep,park2019relational,zhao2022decoupled}. Park \etal transfers mutual relations between different samples rather than the output of individual samples~\cite{park2019relational}. Yu \etal successfully applied knowledge distillation in metric learning~\cite{yu2019learning}. Besides the predicted logits, many works have been proposed to guide the student by the intermediate representations of the teacher model~\cite{passalis2018learning,kim2018paraphrasing,ahn2019variational,koratana2019lit,guan2020differentiable,yue2020matching,zhang2020task,liu2021exploring,shang2021lipschitz,chen2021wasserstein,chen2021distilling,zhu2021complementary,lin2022knowledge,guo2023semantic}. The fundamental problem for distilling intermediate representations is that the dimension of the feature map is different between the teacher and student models. Some previous works~\cite{romero2014fitnets,heo2019comprehensive,yim2017gift,koratana2019lit} overcome these obstacles by building an adaptive module between hidden layers of teacher and student.
Sergey \etal distills intermediate knowledge via matching the attention maps between the teacher and student~\cite{Zagoruyko2017AT}. Similarity Preserving (SP)~\cite{tung2019similarity} unified the dimension at the mini-batch level by matrix operation. Some existing methods re-designed the loss function, including activation transfer loss with boundaries formed by hidden neurons~\cite{heo2019knowledge}, Jacobian~\cite{srinivas2018knowledge}, instance graph~\cite{liu2019knowledge}. Mirzadeh et al.~\cite{mirzadeh2020improved} introduces a multi-step distillation approach by incorporating multiple assistant networks. However, in their method, the teacher, student, and all assistant networks are of homogeneous architectures, meaning they share a similar structural design or configuration.

\noindent \textbf{Knowledge distillation in object detection:}
In recent days, knowledge distillation has been applied into more complex tasks, such as object detection~\cite{li2017mimicking,chen2017learning,wei2018quantization,wang2019distilling,zhengCvpr22LocaliDistill,23PAMI-LocDistill} and semantic segmentation~\cite{liu2019structured,liu2020learning,yang2022cross,ji2022structural,fang2022incremental,fang2023reliable}, \etc. In this paper, we also evaluate our proposed pixel distillation method in object detection. Different from image classification, prediction of object detection contains more complex information. A common way is to distill knowledge from intermediate features of object detector~\cite{chen2017learning,li2017mimicking}. Knowledge from the Feature Pyramid Networks (FPN)~\cite{lin2017feature} can also be used for distillation~\cite{cao2022pkd}. Recently, some works distill knowledge from the detection head of teacher object detector~\cite{zhengCvpr22LocaliDistill,23PAMI-LocDistill,li2022knowledge,wang2023crosskd,lao2023unikd}, which is more complex and efficient.
ScaleKD~\cite{Zhu_2023_CVPR} can distill knowledge between object detectors of different input resolutions, but the teacher and student are of the same architecture. UniKD~\cite{lao2023unikd} can transfer the knowledge in heterogeneous teacher-student pairs, but does not take input resolution into consideration.

\begin{figure*}[!t]
	\begin{center}
		\includegraphics[width=1\linewidth]{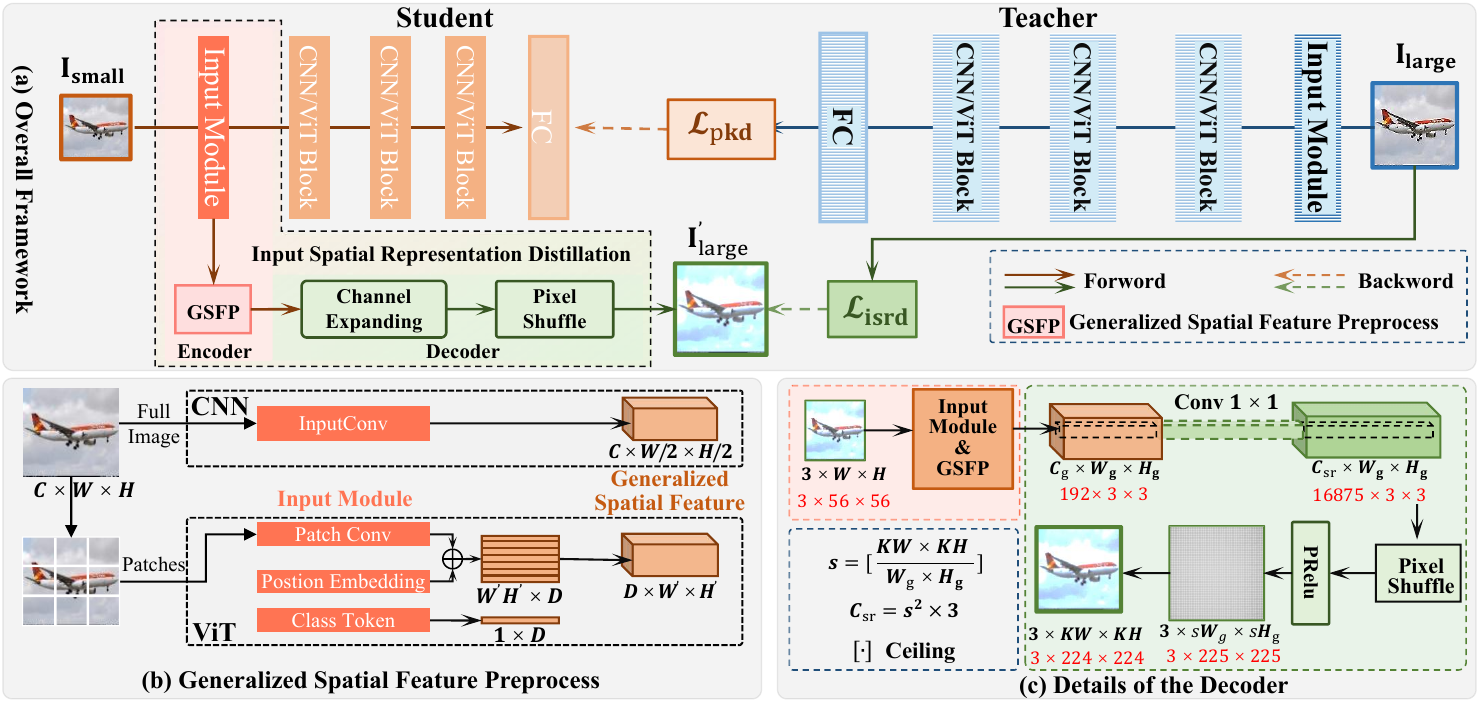}
	\end{center}
	\caption{Illustrations of the vanilla PD. \textbf{(a)} vanilla PD consists of a prediction distillation with an input spatial representation distillation (ISRD). ISRD aims to distill spatial information from the large images to train the input module of the student model. \textbf{(b)} The generalized spatial feature preprocess (GSFP) can transfer spatial features from CNNs and ViTs into the same form. \textbf{(c)} Details of the decoder of the ISRD, red text is an example when the backbone is ViT-Ti/16 and the input size of the student is $56\times 56$.}
	\label{fig_ird}
\end{figure*}

As we illustrated in Table~\ref{tab:tasks}, classical knowledge distillation methods usually use the same input size for teacher and student, the efficiency improvement it can bring is limited to the reduction of model architecture. \textit{However, the use of current distillation approaches is limited when the input size is different between the teacher model and student model}: 
\begin{enumerate}[label=\arabic*)]
        \item Prediction-based distillation methods have high input adaptability for the image classification task because the dimension of the output space is constant. However, using prediction-based distillation in object detection tasks is difficult because of changes in input resolution in inconsistent output dimensions, such as the number of proposals or anchors.
	\item Although both the teacher and student belong to CNNs, some feature-distillation approaches cannot be directly used when the student uses smaller input~\cite{Zagoruyko2017AT,kim2018paraphrasing}. 
	\item Difference in the input level leads to a larger gap between teacher and student, which results in performance degradation for some feature-based distillation methods~\cite{tung2019similarity,liu2021exploring}. 
	\item Due to the structural difference between CNNs and ViTs, most feature-based distillation methods can not be used where the teacher and student belong to CNNs and ViTs, respectively. To be specific, intermediate attention maps from ViT consist of a class token and several patch tokens, where the class token is used to learn class-specific information, and patch tokens are used to learn class-agnostic information. Meanwhile, an intermediate feature from CNN learns semantic-aware information. Therefore, the attention maps from ViT and the intermediate feature from CNN are not in the same form, which makes it difficult to use feature-based distillation methods.
\end{enumerate}
Therefore, a novel distillation method is needed when the compression also occurs at the input level.

\subsection{Fine-tuning and Low-resolution Recognition}

\textbf{Fine-tuning. One way to improve small image recognition is to use fine-tuning (FT) strategy, which adapts a model trained on large images to small images. However, fine-tuning has some limitations: it requires the same model architecture for both source and target models, and it usually only provides a small performance boost for the target model.}

\noindent\textbf{Low-resolution Recognition.}
Low-resolution (LR) recognition aims to achieve high performance with only LR input available in the inference process. In recent years, some works introduce knowledge distillation into the LR image recognition (\ie, \textbf{LR-KD} in Table~\ref{tab:setting})~\cite{li2017perceptual,ge2018low,wang2019improved,qi2021multi,shin2022teaching,ge2020efficient}. However, most methods do not focus on reducing the cost as their main purpose is to achieve better performance. For example, some of them up-sample the LR images to the same size as the HR images to use the existing distillation algorithm more conveniently~\cite{wang2019improved,zhu2019low}, while others use the same architecture as the teacher model~\cite{ge2020efficient,qi2021multi,chen2022super,shin2022teaching, huang2022feature,hu2022cross}. Furthermore, some works modify the network architecture by using fewer pooling layers or extra modules~\cite{ge2018low,li2017perceptual}, which makes it difficult to use large-scale pre-training models and reduces their generalization ability. As a result, these methods only perform well on small-scale databases or simple scenarios like face recognition. Taking the latest work in the field of LR face recognition---FMD~\cite{huang2022feature}---as an example, FMD aims to enhance the performance of students with small input by using a teacher with large input, while maintaining the same architecture for the student. However, the FMD configuration restricts its use to situations where the input size of the teacher and the student are limited to $92\times 92$ and $44\times 44$, respectively. As a result, FMD fails to meet the criteria of either input adaptability or network adaptability.

Previous LR recognition methods, as discussed in~\cite{ge2018low,li2017perceptual,chen2022super,huang2022feature}, are limited in their applicability to specific input size settings or network architectures due to their complex designs. In contrast, our pixel distillation paradigm offers a more flexible approach. By generalizing classical knowledge distillation in the input level, pixel distillation provides more options for deployment and can be adapted to various input sizes and network architectures for both the teacher and student models. This adaptability sets pixel distillation apart from previous LR recognition methods, which suffer from a lack of input and network adaptability.


\section{Pixel Distillation}
In this section, our goal is to address the pixel-level distillation problem within the image classification and object detection tasks. Initially, we will present a foundational overview of how knowledge distillation is applied to image classification. Then, we propose a straightforward pixel distillation baseline for image classification that we have developed using our novel input spatial representation distillation module. Subsequently, we explore integrating feature-level distillation with pixel distillation, in which we utilize the teacher-assistant-student framework. Finally, we adapt pixel distillation to object detection, and introduce an aligned feature preservation strategy for the assistant network, to tackle the challenge of inconsistent output dimensions caused by varied image resolutions.

\subsection{Preliminary of KD in Image Classification}
Traditional knowledge distillation approaches train a student model by learning the information from the teacher models. Based on the way to obtain the supervision information, we classify previous works in image classification into two categories: prediction-based methods and feature-based methods. 

\noindent\textbf{Prediction-based distillation methods} train the student by using the class scores predicted by the teacher. One essential way is to regard the predicted logits of the teacher model as the soft target of the student \cite{ba2014deep,hinton2015distilling}. The loss is:
\begin{equation}
	\mathcal{L}_{\rm pkd}(\mathbf{y},\mathbf{x}_{\rm t},\mathbf{x}_{\rm s}) = (1-\alpha)\mathcal{L}_{\rm cls}(\mathbf{y}, \mathbf{x}_{\rm s}) + {\alpha}T^{2}{\mathcal{L}_{\rm kl}}(\mathbf{p}_{\rm t}, \mathbf{p}_{\rm s}),
	\label{eq:pkd}
\end{equation}
where $\mathbf{p}_{\rm t}={\rm softmax}(\frac{\mathbf{x}_{\rm t}}{T})$ is the class scores predicted by the teacher, $\mathbf{y}$ is the ground truth, $\mathbf{p}_{\rm s}= {\rm log} ({\rm softmax}(\frac{\mathbf{x}_{\rm s}}{T}))$, $\mathbf{x}_{\rm t}$ and $\mathbf{x}_{\rm s}$ are the predicted class scores of the teacher and student model, respectively, $T$ is a temperature parameter, $\alpha$ is a hyperparameter to balance the classification loss $L_{\rm cls}$ and the Kullback–Leibler divergence loss $L_{\rm kl}$.

\noindent\textbf{Feature-based distillation methods} guide the student model using intermediate representations from the teacher model. The learning process of these methods can be expressed as follows:
\begin{equation}
	\mathcal{L}_{\rm fkd}(\mathcal{F}_{\rm t},\mathcal{F}_{\rm s},\mathbf{x}_{\rm s}) = \mathcal{L}_{\rm cls}(\mathbf{y}, \mathbf{x}_{\rm s}) + {\beta}\sum_{i\in \mathbf{B}}{\delta}(g_{\rm t}(\mathbf{F}_{{\rm t},i}), g_{\rm s}(\mathbf{F}_{{\rm s},i})),
	\label{eq:fkd}
\end{equation}
where $\mathcal{F}_{\rm s}=\{\mathbf{F}_{{\rm s},1},\mathbf{F}_{{\rm s},2},...,\mathbf{F}_{{\rm s},M}\}$ represents the intermediate features of student, while $\mathcal{F}_{\rm t}=\{\mathbf{F}_{{\rm t},1},\mathbf{F}_{{\rm t},2},...,\mathbf{F}_{{\rm t},M}\}$ denotes the features of teacher. The variable $M$ represents the number of blocks in the network. $\mathbf{B}$ is the set of selected features, which varies for different methods. $g_{\rm t}(\cdot)$ and $g_{\rm s}(\cdot)$ are functions to extract information from intermediate features. $\delta(\cdot)$ is the distance metric function. $\beta$ is a hyperparameter to balance the classification loss and feature distillation loss.

\begin{figure}[!t]
	\centering
	\begin{subfigure}{0.48\linewidth}
		\includegraphics[width=1\linewidth]{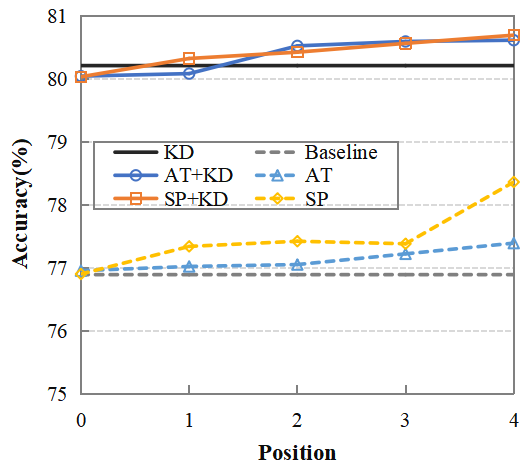}
		\label{fig:position_k1}
		\vspace{-0.3cm}
		\caption{Knowledge distillation}
	\end{subfigure}
	\hspace{0.1cm}
	\begin{subfigure}{0.48\linewidth}
		\includegraphics[width=1\linewidth]{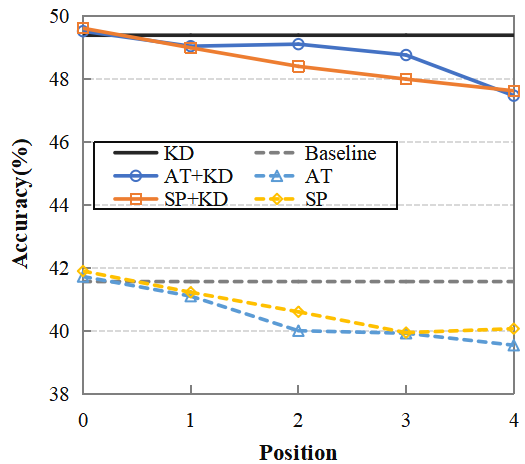}
		\label{fig:position_k2}
		\vspace{-0.3cm}
		\caption{Pixel distillation}
	\end{subfigure}
	\caption{Study about the distillation position in traditional knowledge distillation ($K$=1) and our pixel distillation ($K$=4) in image classification. The teacher is ResNet50 with $224\times 224$ input, and the student is ResNet18 with $\frac{224}{K}\times \frac{224}{K}$ input.}
	\label{fig:position}
\end{figure}

\subsection{Building A Simple Baseline}
In this paper, our objective is to train the student model with the help of the teacher model, where both the network architectures and the input size are different. The teacher model takes large images and utilizes heavy networks, while the student model takes small images as input and uses a lightweight network. To identify the best cost scheme, a pixel distillation method should adhere to two criteria: Firstly, it should be applicable to various architectures of both the teacher and student models, including different CNNs and ViTs. Secondly, the method should be adaptable to different small input sizes of the student. In this section, we introduce a simple one-stage trained baseline method vanilla PD that satisfies these two criteria.

\subsubsection{Framework of vanilla PD}
As shown in Fig.~\ref{fig_ird}a, the proposed baseline consists of two distillation processes: 
\begin{enumerate}[label=\arabic*)]
	\item Following prediction-based distillation methods~\cite{hinton2015distilling,zhao2022decoupled}, we use the logits of the teacher as a part of the supervision for the student. Prediction-based distillation methods are unaffected by network architecture and input, so they can naturally satisfy the abovementioned criteria.
	\item An Input Spatial Representation Distillation (ISRD) mechanism is proposed to let the input module of the student learn valuable spatial knowledge from the large input of the teacher. ISRD needs to be designed carefully to satisfy the two criteria.
\end{enumerate}

The loss of vanilla PD is defined as:
\begin{equation}
	\mathcal{L}_{\text {vanilla}}= \mathcal{L}_{\text {pkd}} + \gamma\mathcal{L}_{\text {isrd}},
\end{equation}
where $\mathcal{L}_{\text {pkd}}$ is the loss of the prediction-based distillation, and $\mathcal{L}_{\text {isrd}}$ is the loss of the ISRD. $\gamma$ is a hyperparameter to balance $\mathcal{L}_{\text {pkd}}$ and $\mathcal{L}_{\text {isrd}}$.

\begin{algorithm}[!t]
	\caption{Input Spatial Representation Distillation}
	\begin{algorithmic}[1]
		\Require Input of the student model $\mathbf{I}_{{\rm lr}} \in \mathbb{R}^{3\times W\times H}$, Input of the teacher model $\mathbf{I}_{{\rm hr}} \in \mathbb{R}^{3\times KW\times KH}$
		\If{stuent is CNN}
		\State Directly use the input feature of the student model: $\mathbf{F}_{{\rm g},0}=\mathbf{F}_{{\rm s},0}, \in \mathbb{R}^{C \times \frac{W}{2} \times \frac{H}{2}}$;
		\ElsIf{stuent is ViT}
		\State Extract patch embedding: $\mathbf{F}_{{\rm patch},0} \in \mathbb{R}^{W^{'}H^{'} \times D}$;
		\State Transform the form of patch embedding, obtain $\mathbf{F}_{{\rm g},0} \in \mathbb{R}^{D\times W^{'}\times H^{'}}$;
		\EndIf
		\State Suppose the generalized input spatial feature as $\mathbf{F}_{{\rm g},0} \in \mathbb{R}^{C_{\rm g}\times W_{\rm g}\times H_{\rm g}}$;
		\State Calculate scale factor: $s = [\frac{KW\times KH}{W_{\rm g}\times H_{\rm g}}]$;
		\State Calculate expanded channel number: $C_{\rm sr} = 3s^2$; 
		\State Expand feature by a $1\times 1$ convolutional layer, obtain $\mathbf{F}_{{\rm sr}, 0} \in \mathbb{R}^{C_{\rm sr}\times W_{\rm g}\times H_{\rm g}}$;
		\State Transform the expanded feature into 3 channels by the pixel shuffle operation, obtain $\mathbf{I}_{{\rm sr}, 0} \in \mathbb{R}^{3\times sW_{\rm g}\times sH_{\rm g}}$;
		\State Obtain the pseudo large image $\mathbf{I}_{\rm hr}^{'} \in \mathbb{R}^{3\times KW\times KH}$ by a crop operation;
		\Ensure $\mathbf{I}_{\rm hr}^{'}$
	\end{algorithmic}
	\label {alg:isrd}
\end{algorithm}

\subsubsection{Input Spatial Representation Distillation}
Based on Fig.\ref{fig:position}, it can be observed that the knowledge of the input convolution layer is more beneficial than that of the intermediate layers in pixel distillation. This is because the change of input image leads to larger gaps between intermediate features of the student and teacher, which will make it difficult for the student model to imitate the teacher model~\cite{cho2019efficacy}. Considering the small input in pixel distillation would lead to inadequate information on the shallow features of the student, the proposed ISRD is used just after the input convolutions of CNN or ViT. As shown in Fig.~\ref{fig_ird}a, the ISRD is an autoencoder that takes the input feature of the student as the input and outputs the large image. The encoder of ISRD transforms the spatial information of CNN and ViT into the same form, and the decoder of ISRD predicts the large image by the transformed feature. In this paper, we calculate the $l_1$ loss between the pseudo large image $\mathbf{I}_{\rm hr}^{'}$ and real large image $\mathbf{I}_{\rm hr}$ as the loss for ISRD module. Suppose the input of the student is an image with a width of $W$ pixels and a height of $H$ pixels, and the input of the teacher is an image with a width of $KW$ pixels and a height of $KH$ pixels, the loss of ISRD is defined as:
\begin{equation}
	\mathcal{L}_{\text {isrd}} = \frac{1}{3\times KW\times KH} { \parallel \mathbf{I}_{\rm hr}^{'} - \mathbf{I}_{\rm hr} \parallel}_{1}^{1},
\end{equation}
we provide the detailed learning process of the ISRD in Algorithm~\ref{alg:isrd}.

\noindent \textbf{Encoder of ISRD.}
The ISRD encoder is composed of a student input convolution layer and a GSFP operation. As we illustrated in Fig.~\ref{fig_ird}b, the GSFP operation is a parameter-free operation, the student input convolution layer is the encoder's only learnable parameter. Both CNN and ViT use a convolution layer to map the input images into feature space, but the form of their input features are very different, so we need to transform the features of CNN and ViT into the same form to achieve a generalized distillation. As shown in Fig.~\ref{fig_ird}b, given a small input $\mathbf{I}_{{\rm lr}} \in \mathbb{R}^{3\times W\times H}$, the feature map generated by the input module of CNN usually is $\mathbf{F}_{{\rm s},0}, \in \mathbb{R}^{C \times \frac{W}{2} \times \frac{H}{2}}$, where $C$ is the channel number. Different from CNN, ViT splits the input image into patches and its input feature contains a series of patch tokens and one class token. The patch tokens $\mathbf{F}_{{\rm patch},0}, \in \mathbb{R}^{N \times D}$ are derived from the summation of patch features and position embeddings, which contain the spatial information of the input, where $N=W^{'}\times H^{'}$ is the number of patches and $D$ is the hidden size of the tokens. In this paper we transform patch tokens into the size of $D\times W^{'} \times H^{'}$ to make it has the same form as the input feature of the CNN. For both CNN and ViT, We regard the generalized input spatial feature as $\mathbf{F}_{{\rm g},0} \in \mathbb{R}^{C_{\rm g}\times W_{\rm g}\times H_{\rm g}}$.

\noindent \textbf{Decoder of ISRD.}
As shown in Fig.~\ref{fig_ird}c, given the generalized input spatial feature $\mathbf{F}_{{\rm g}}$, we need to expand its volume the same as the large input $\mathbf{I}_{{\rm hr}} \in \mathbb{R}^{3\times KW\times KH}$. In this paper, we use a $1\times 1$ convolutional layer to achieve channel expansion. To be specific, the scale factor $s$ of the spatial size should be $\frac{KW\times KH}{W_{\rm g}\times H_{\rm g}}$, but in most cases, $s$ is not an integer, so we use the round-up operation for $s$ before obtaining the expanded channel number:
\begin{equation}
	s = [\frac{KW\times KH}{W_{\rm g}\times H_{\rm g}}], \quad C_{\rm sr} = 3s^2,
	\label{eq:channel}
\end{equation}
where $C_{\rm sr}$ is the channel number of the extended feature, $[\cdot]$ is the round-up operation, the expanded input spatial feature is $\mathbf{F}_{{\rm sr}, 0} \in \mathbb{R}^{C_{\rm sr}\times W_{\rm g}\times H_{\rm g}}$. Then, we use a pixel shuffle operation~\cite{shi2016real} to transform the extended feature into a feature map of 3 channels and obtain $\mathbf{I}_{{\rm sr}, 0} \in \mathbb{R}^{3\times sW_{\rm g}\times sH_{\rm g}}$. Since we use the round-up operation when calculating the scale factor $s$, this map may be slightly larger than the large image, so we use a crop operation to obtain the final pseudo large image $\mathbf{I}_{\rm hr}^{'}$.

\begin{figure}[!t]
	\begin{center}
		\includegraphics[width=0.85\linewidth]{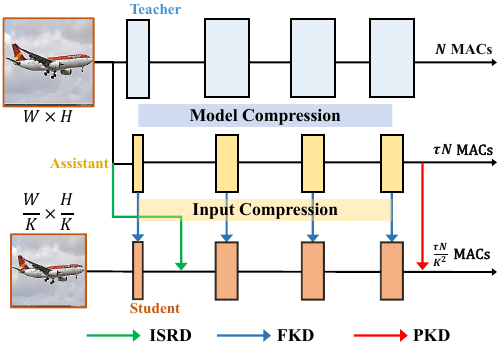}
	\end{center}
	\caption{Illustrations of the Teacher-Assistant-Student (TAS) framework for image classification task. The assistant model splits the pixel distillation into two stages: the model compression stage which reduces the computational cost by a factor of $\tau$ ($0<\tau<1$) by using lightweight network architecture, and the input compression stage further reduces the input size and computational cost by a factor of $\frac{1}{K^2}$ by using small input.}
	\label{fig_tas}
\end{figure}

\subsection{Teacher-Assistant-Student Framework.}
In the classical teacher-student framework of knowledge distillation scheme, the teacher and student have the same input size and different network architecture. However, in the pixel distillation scheme, the teacher and student have different input sizes and network architecture, which makes it more difficult for the student to successfully mimic the teacher~\cite{cho2019efficacy}. To reduce the learning difficulty for the student in pixel distillation, this paper introduces an assistant network into the classical teacher-student framework to decouple the process of pixel distillation into the model compression stage and input compression stage. As shown in Fig.~\ref{fig_tas}, the assistant network maintains the same large input size as the teacher model and has the same lightweight network architecture as the student. In the model compression stage, the assistant model is regarded as a student model to receive knowledge from the teacher model which has complex architecture. In the input compression stage, the assistant model is regarded as a teacher model to transfer spatial knowledge from the large input into the student model. The whole framework is called the Teacher-Assistant-Student (TAS) framework.

The proposed TAS framework will bring three-fold benefits for pixel distillation:
\begin{enumerate}[label=\arabic*)]
	\item TAS is a two-stage learning framework and the learning difficulty of every single stage is smaller than the one-stage teacher-student framework, which will bring a higher performance gain for the student model.
	\item Off-the-shelf knowledge distillation methods can be used in the model compression stage of the TAS framework.
	\item In the input compression stage, since the assistant model and student model have the same network architecture, their features are of the same form and number, only the number of channels (CNN) or patches (ViT) is different, which makes it much easier to use feature distillation strategies.
\end{enumerate}

In this paper, we design a simple feature distillation strategy that applies an upsampling operation to features from the assistant model. Suppose $\mathcal{F}_{\rm s}=\{\mathbf{F}_{{\rm s},1},\mathbf{F}_{{\rm s},2},...,\mathbf{F}_{{\rm s},M}\}$ is the spatial features of a student model, $\mathcal{F}_{\rm a}=\{\mathbf{F}_{{\rm a},1},\mathbf{F}_{{\rm a},2},...,\mathbf{F}_{{\rm a},M}\}$ is the spatial feature of the assistant model, $M$ is the number of blocks in the network, the feature distillation loss of the student in the input compression stage is:
\begin{equation}
	\mathcal{L}_{\rm icf}(\mathcal{F}_{\rm a},\mathcal{F}_{\rm s}) = \sum_{i\in \mathbf{B}}{\delta}({\rm UP}(\mathbf{F}_{{\rm s},i}), \mathbf{F}_{{\rm a},i}),
	\label{eq:icfkd}
\end{equation}
where $\mathbf{B}$ denotes the set of selected features. ``UP'' denotes the upsampling operation that is used to make the spatial size of the student feature the same as that of the assistant feature. To be specific, for feature maps from CNN, we directly upsample the feature map in the spatial dimension. For attention maps from ViT, we first expand the spatial dimension into two dimensions, and then apply the upsampling operation. The overall loss of the input compression in the TAS is:
\begin{equation}
	\mathcal{L}_{\rm ic} = \mathcal{L}_{\rm pkd}(\mathbf{y},\mathbf{x}_{\rm a},\mathbf{x}_{\rm s}) + \gamma\mathcal{L}_{\text {isrd}} + \eta\mathcal{L}_{\rm icf}(\mathcal{F}_{\rm a},\mathcal{F}_{\rm s}),
	\label{eq:ic}
\end{equation}
where $\eta$ is the loss weight for the feature distillation loss.

\subsection{Pixel Distillation in Object Detection}

\begin{figure}[!t]
	\begin{center}
		\includegraphics[width=1\linewidth]{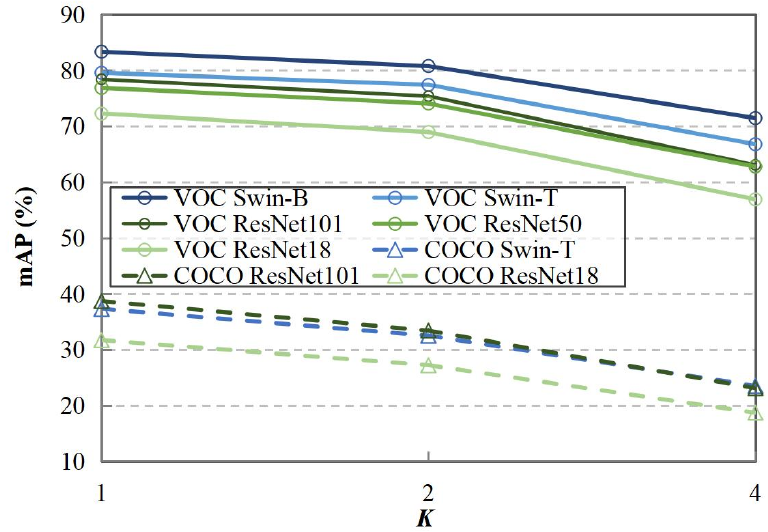}
	\end{center}
	\caption{Analysis of how input resolution affects the performance of object detection. We report the mAP (\%) of RetinaNet~\cite{lin2017focal} on the PASCAL VOC and COCO datasets.}
	\label{fig:odsize}
\end{figure}

In this paper, we also extend our pixel distillation paradigm into the fields of object detection. We first analyze how the image resolution affects the performance of the object detectors and the distillation process. Then, we design an aligned feature for preservation (AFP) strategy to align the output dimensions of detectors at each stage. 

\noindent \textbf{Effect of input resolution:} As shown in Fig.~\ref{fig:odsize}, we first analyze how the performance of the object detector is affected by the input resolution. On the PASCAL VOC and COCO datasets, We report the mAP (\%) of RetinaNet~\cite{lin2017focal} under several backbones. We observe that with a 4x reduction in resolution ($K$=4), the mAP for all examined models drops by at least 10\%, demonstrating that input resolution significantly influences the effectiveness of object detectors. Furthermore, decreasing input resolution not only affects model performance but also alters output characteristics, such as the number of anchor boxes, thereby complicating the process of transferring knowledge from the teacher’s detection head during distillation.

\begin{figure}[!t]
	\begin{center}
		\includegraphics[width=1\linewidth]{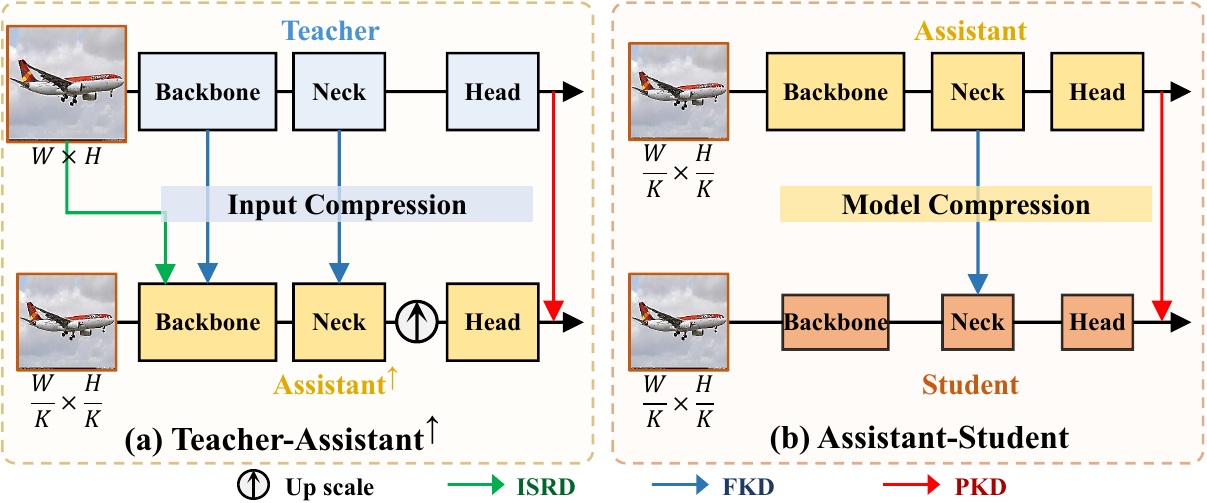}
	\end{center}
	\caption{The overall teacher-assistant-student framework for object detection involves employing the assistant network's features in distinct manners across two stages to align the output dimensions. `Assistant$^\uparrow$' denotes that the features of the assistant are upscaled to match the dimensionality of the teacher's features.}
	\label{fig_odtas}
\end{figure}

\noindent \textbf{Aligned feature for preservation:} As aforementioned before, decreasing input resolution will affect the output characteristics such as the number of anchor boxes, which makes it hard to perform pixel distillation in one stage if we want to preserve knowledge from the prediction head of the teacher. To address this issue, we have refined the two-stage teacher-assistant-student framework in this study. Fig.~\ref{fig_odtas} illustrates that since the assistant network is not utilized during inference, its architecture can be adapted to counteract the effects of resolution reduction on the detector. Specifically, in the first input compression stage, both the teacher and assistant use the same backbone network. The assistant's features are upsampled to match the spatial resolution of the teacher's features, which we refer to as `Assistant$^\uparrow$'. This approach allows the assistant detector to utilize the same anchoring schema as the teacher, facilitating the preservation of knowledge at the prediction level. The associated loss function for this object detection input compression stage is detailed as follows:
\begin{equation}
	\mathcal{L}_{\rm od, ic} = \mathcal{L}_{\rm pkd} + \mathcal{L}_{\rm fkd} + \gamma\mathcal{L}_{\text {isrd}}
	\label{eq:icod}
\end{equation}
where $\mathcal{L}_{\rm pkd}$ and $\mathcal{L}_{\rm fkd}$ denote the loss for the prediction distillation and feature distillation, respectively.

In the model compression phase, which is the second stage, both the teacher and student are provided with the same input, which allows for the removal of the feature upsampling operation within the assistant detector. Consequently, the assistant now adopts the same anchoring schema as the student. This alignment makes it possible to employ standard knowledge distillation techniques.

To sum up, by effectively manipulating the features and anchor configurations of the assistant network, our pixel distillation method is extended to the object detection task. Also, both prediction-based and feature-based distillation techniques can be used in each of the two stages. It is important to note that the input compression stage alters the structure of the assistant network and brings additional computational costs. Therefore, the sequence of the two stages in the Teacher-Assistant-Student (TAS) framework is not interchangeable for object detection tasks. The input compression stage must precede the model compression stage to ensure that the assistant's architecture is correctly configured for each stage.

\begin{table*}[!t]
	\centering
	\scriptsize
	\caption{The configuration of teacher and student models for six experiment settings. We use different colors to present settings related \textbf{only to input
	} (Stg/Trans., Stg/Trans Red.), \textbf{only to architecture} (Param.), and \textbf{to both input and architecture} (MACs, Compute Red.).}
	\resizebox{1\linewidth}{!}{
		\begin{tabular}{c|c|lc|lcccccc}
			\hline
			\multicolumn{1}{r}{} & \multicolumn{1}{r}{} &       &       &       & \textbf{(a)} & \textbf{(b)} & \textbf{(c)} & \textbf{(d)} & \textbf{(e)} & \textbf{(f)} \\
			\hline
			\multirow{3}[4]{*}{\begin{turn}{88}\textbf{Teacher}\end{turn}} & \multicolumn{3}{c|}{\multirow{2}[2]{*}{\diagbox{$\qquad$\textbf{\textcolor[rgb]{ 1,  .663,  .663}{\textbf{Input↓}}}$\quad\qquad$}{\textbf{\textcolor[rgb]{ .663,  .816,  .557}{Architecture→}}}}} & \cellcolor[rgb]{ .886,  .937,  .855}Network & \cellcolor[rgb]{ .886,  .937,  .855}ResNeXt50 & \cellcolor[rgb]{ .886,  .937,  .855}ResNet50 & \cellcolor[rgb]{ .886,  .937,  .855}ResNet18 & \cellcolor[rgb]{ .886,  .937,  .855}ResNet50 & \cellcolor[rgb]{ .886,  .937,  .855}ViT-B/16 & \cellcolor[rgb]{ .886,  .937,  .855}ViT-B/16 \\
			& \multicolumn{3}{c|}{} & \cellcolor[rgb]{ .886,  .937,  .855}Param. (M) & \cellcolor[rgb]{ .886,  .937,  .855}25.03  & \cellcolor[rgb]{ .886,  .937,  .855}25.56  & \cellcolor[rgb]{ .886,  .937,  .855}11.69  & \cellcolor[rgb]{ .886,  .937,  .855}25.56  & \cellcolor[rgb]{ .886,  .937,  .855}86.57  & \cellcolor[rgb]{ .886,  .937,  .855}86.57  \\
			\cmidrule{2-11}          & \multicolumn{1}{l|}{\cellcolor[rgb]{ 1,  .882,  .882}Size $224^2$} & \cellcolor[rgb]{ 1,  .882,  .882}Stg/Trans.(KB) & \cellcolor[rgb]{ 1,  .882,  .882}147 & \cellcolor[rgb]{ .867,  .922,  .969}MACs (G) & \cellcolor[rgb]{ .867,  .922,  .969}4.260  & \cellcolor[rgb]{ .867,  .922,  .969}4.110  & \cellcolor[rgb]{ .867,  .922,  .969}1.820  & \cellcolor[rgb]{ .867,  .922,  .969}4.110  & \cellcolor[rgb]{ .867,  .922,  .969}16.850  & \cellcolor[rgb]{ .867,  .922,  .969}16.850  \\
			\hline
			\hline
			\multirow{6}[6]{*}{\begin{sideways}\textbf{Student}\end{sideways}} & \multicolumn{3}{c|}{\multirow{2}[2]{*}{\diagbox{$\qquad$\textbf{\textcolor[rgb]{ 1,  .663,  .663}{\textbf{Input↓}}}$\quad\qquad$}{\textbf{\textcolor[rgb]{ .663,  .816,  .557}{Architecture→}}}}} & \cellcolor[rgb]{ .886,  .937,  .855}Network & \cellcolor[rgb]{ .886,  .937,  .855}ResNet34 & \cellcolor[rgb]{ .886,  .937,  .855}ResNet18 & \cellcolor[rgb]{ .886,  .937,  .855}ShuffleNetV2 1.0 & \cellcolor[rgb]{ .886,  .937,  .855}ViT-Ti/16 & \cellcolor[rgb]{ .886,  .937,  .855}ResNet18 & \cellcolor[rgb]{ .886,  .937,  .855}ViT-Ti/16 \\
			& \multicolumn{3}{c|}{} & \cellcolor[rgb]{ .886,  .937,  .855}Param. (M) & \cellcolor[rgb]{ .886,  .937,  .855}21.80  & \cellcolor[rgb]{ .886,  .937,  .855}11.69  & \cellcolor[rgb]{ .886,  .937,  .855}2.28  & \cellcolor[rgb]{ .886,  .937,  .855}5.69  & \cellcolor[rgb]{ .886,  .937,  .855}11.69  & \cellcolor[rgb]{ .886,  .937,  .855}5.69  \\
			\cmidrule{2-11}          & \cellcolor[rgb]{ 1,  .882,  .882}K=2 & \multicolumn{1}{p{6.75em}}{\cellcolor[rgb]{ 1,  .882,  .882}Stg/Trans.(KB)} & \cellcolor[rgb]{ 1,  .882,  .882}36.75 & \cellcolor[rgb]{ .867,  .922,  .969}MACs (G) & \cellcolor[rgb]{ .867,  .922,  .969}0.967  & \cellcolor[rgb]{ .867,  .922,  .969}0.487  & \cellcolor[rgb]{ .867,  .922,  .969}0.041  & \cellcolor[rgb]{ .867,  .922,  .969}0.273  & \cellcolor[rgb]{ .867,  .922,  .969}0.487  & \cellcolor[rgb]{ .867,  .922,  .969}0.273  \\
			& \cellcolor[rgb]{ 1,  .882,  .882}Size $112^2$ & \multicolumn{1}{p{6.75em}}{\cellcolor[rgb]{ 1,  .882,  .882}Stg/Trans Red.} & \cellcolor[rgb]{ 1,  .882,  .882}75.00\% & \cellcolor[rgb]{ .867,  .922,  .969}Comput Red. & \cellcolor[rgb]{ .867,  .922,  .969}77.29\% & \cellcolor[rgb]{ .867,  .922,  .969}88.16\% & \cellcolor[rgb]{ .867,  .922,  .969}97.72\% & \cellcolor[rgb]{ .867,  .922,  .969}93.36\% & \cellcolor[rgb]{ .867,  .922,  .969}97.11\% & \cellcolor[rgb]{ .867,  .922,  .969}98.38\% \\
			\cmidrule{2-11}          & \cellcolor[rgb]{ 1,  .882,  .882}K=4 & \cellcolor[rgb]{ 1,  .882,  .882}Stg/Trans.(KB) & \cellcolor[rgb]{ 1,  .882,  .882}9.1875 & \cellcolor[rgb]{ .867,  .922,  .969}MACs (G) & \cellcolor[rgb]{ .867,  .922,  .969}0.268  & \cellcolor[rgb]{ .867,  .922,  .969}0.130  & \cellcolor[rgb]{ .867,  .922,  .969}0.012  & \cellcolor[rgb]{ .867,  .922,  .969}0.055  & \cellcolor[rgb]{ .867,  .922,  .969}0.130  & \cellcolor[rgb]{ .867,  .922,  .969}0.055  \\
			& \cellcolor[rgb]{ 1,  .882,  .882}Size $56^2$ & \cellcolor[rgb]{ 1,  .882,  .882}Stg/Trans Red. & \cellcolor[rgb]{ 1,  .882,  .882}93.75\% & \cellcolor[rgb]{ .867,  .922,  .969}Comput Red. & \cellcolor[rgb]{ .867,  .922,  .969}93.71\% & \cellcolor[rgb]{ .867,  .922,  .969}96.84\% & \cellcolor[rgb]{ .867,  .922,  .969}99.33\% & \cellcolor[rgb]{ .867,  .922,  .969}98.67\% & \cellcolor[rgb]{ .867,  .922,  .969}99.23\% & \cellcolor[rgb]{ .867,  .922,  .969}99.68\% \\
			\hline
		\end{tabular}%
	}
	\label{tab:setting}%
\end{table*}%

\section{Experiments}

\subsection{Settings}

\noindent\textbf{Datasets.} To evaluate the performance of our proposed method in realistic settings, we choose three widely used datasets that reflect different challenges and characteristics of image classification tasks, \ie, CUB (Caltech-UCSD Birds-200-2011)~\cite{wah2011caltech}, Aircraft (FGCV-aircraft-2013b)~\cite{maji2013fine}, and ImageNet~\cite{deng2009imagenet}. CUB is a fine-grained dataset that consists of 200 categories of birds, there are 5,994 training images and 5,794 testing images. Aircraft contains 100 categories of aircraft and each category has 100 images. The train, validation, and test set have 3,334, 3,333, and 3,333 images, respectively. ImageNet has 1,000 categories, each category contains approximately 1,300 training and 50 validation images per category. In total, it contains 129,395 and 5,000 images for train and validation, respectively.
Besides, to evaluate our proposed method on the object detection task, we select two widely used benchmarks, \ie, PASCAL VOC~\cite{everingham2010pascal} and COCO 2017 dataset~\cite{lin2014microsoft}. For PASCAL VOC, we use 5,000 trainval images in VOC2007 and 16,000 trainval images in VOC2012 for training, and 5,000 test images in VOC 2007 for evaluation. For COCO, we use 115,000 trainval135k images for training, and 5,000 minival set as validation.

\noindent\textbf{Metrics.} To evaluate the classification performance, we use Top-1 accuracy. For object detection, we report mean Average Precision (AP) as an evaluation metric. We also provide several metrics to evaluate how the input size and architecture complexity affect the cost of the student model. For input-related costs, we provide the storage and transmission (Stg/Trans.) for one image. For architecture-related costs, we provide the number of parameters (Param.). Moreover, we use the number of multiply–accumulate operations (MACs) to measure the computational complexity of the model, which is related to both the input size and architecture complexity. 

We also propose a metric to evaluate the reduction of storage and transmission cost (Stg/Trans Red.), which is calculated as: 
\begin{equation}
	\text{Stg/Trans Red.} = 1-\frac{\text{Stg/Trans. (Student)}}{\text{Stg/Trans. (Teacher)}}.
\end{equation}
another metric is proposed to evaluate the reduction of computational complexity (Comput Red.), which is calculated as: 
\begin{equation}
	\text{Comput Red.} = 1-\frac{\text{MACs (Student)}}{\text{MACs (Teacher)}}.
\end{equation}

\noindent\textbf{Implementation details.} 
For the image classification task, we use mini-batch stochastic gradient descent (SGD) as the optimizer, and the momentum and the weight decay are set as 0.9 and 0.0005, respectively. On CUB and Aircraft, we set the learning rate as 0.01, 0.02, and 0.001 for ResNet, ShuffleNetV2, and ViT, respectively. We train the model by 120 epochs with batch size 64, the learning rate is reduced by a factor of 10 after every 30 epochs. On ImageNet, we set the learning rate as 0.001 and keep the remaining setting the same as CUB and Aircraft. The value of $\gamma$ is determined by the architecture of the student network we analyze it in Section~\ref{sec:ablation}. The value of $\eta$ is set as 10. 
For the object detection task, all models are implemented under the MMDetection~\cite{chen2019mmdetection} toolkit.
Our code is implemented on the basis of PyTorch \cite{paszke2017automatic}, and all experiments are carried out on an NVIDIA GeForce 3090 GPU.

\begin{table}[!t]
	\centering
	\footnotesize
	\caption{Results on the CUB dataset for setting (a) to (c). The best is shown in bold. Each experiment is repeated five times and we report the mean value.}
	\resizebox{\linewidth}{!}{
		\begin{tabular}{l|cc|cc|cc}
			\hline
			& \multicolumn{2}{c|}{(a)} & \multicolumn{2}{c|}{(b)} & \multicolumn{2}{c}{(c)} \\
			& $K=2$  & $K=4$  & $K=2$  & $K=4$  & $K=2$  & $K=4$\\
			\hline
			Teacher & \multicolumn{2}{c|}{81.84 } & \multicolumn{2}{c|}{80.46 } & \multicolumn{2}{c}{76.89 } \\
			\hline
			Baseline-FS & 37.54  & 23.73  & 37.16  & 24.52  & 35.67  & 22.36  \\
			Baseline-FT & 65.41  & 43.68  & 63.31  & 41.56  & 63.53  & 43.05  \\
			\hline
			KD    & 70.40  & 49.32  & 69.26  & 49.38  & 65.69  & 44.09  \\
			AT    & 61.84  & 4.63  & 58.53  & 4.55  & 43.61  & 6.97  \\
			AT+KD & 65.25  & 4.64  & 62.74  & 7.20  & 47.28  & 9.00  \\
			SP    & 68.35  & 42.31  & 65.96  & 40.07  & 64.35  & 40.45  \\
			SP+KD & 70.41  & 48.01  & 69.39  & 47.61  & 65.69  & 42.54  \\
			IC    & 67.36  & 44.53  & 66.03  & 46.02  & 62.11  & 42.90  \\
			IC+KD & 69.17  & 49.72  & 69.81  & 48.30  & 66.17  & 44.07  \\
			DKD   & 70.08  & 47.01  & 68.77  & 48.78  & 66.41  & 45.15  \\
			\hline
			vanilla PD (One-Stage) & 70.86  & 50.48  & 69.94  & 50.34  & 66.32  & 44.69  \\
			TAS (Two-Stage) & \textbf{72.59} & \textbf{51.52} & \textbf{71.65} & \textbf{52.44} & \textbf{67.67} & \textbf{45.93} \\
			\hline
		\end{tabular}%
	}
	\label{tab:cub}%
\end{table}%

\subsection{Experimental Results of Image Classification}
In this section, we compared our method with previous knowledge distillation methods including prediction distillation methods KD~\cite{hinton2015distilling}, DKD~\cite{zhao2022decoupled}, feature distillation methods AT~\cite{Zagoruyko2017AT}, SP~\cite{tung2019similarity}, IC~\cite{liu2021exploring}. AT~\cite{Zagoruyko2017AT} is selected as the representative of methods requiring the same spatial size for features from teacher and student, we use average pooling to align feature maps following~\cite{tian2019contrastive}. SP~\cite{tung2019similarity} is the representative of methods that do not require the same spatial size for teacher and student features, which can be directly used to resolve the pixel distillation problem. IC~\cite{liu2021exploring} is the representative of methods that contains an extra adaptive module to align features between teacher and student. We also provide the performance of the combination of each feature distillation method with the prediction distillation method~\cite{hinton2015distilling}.

\begin{figure}[!t]
	\begin{center}
		\includegraphics[width=1\linewidth]{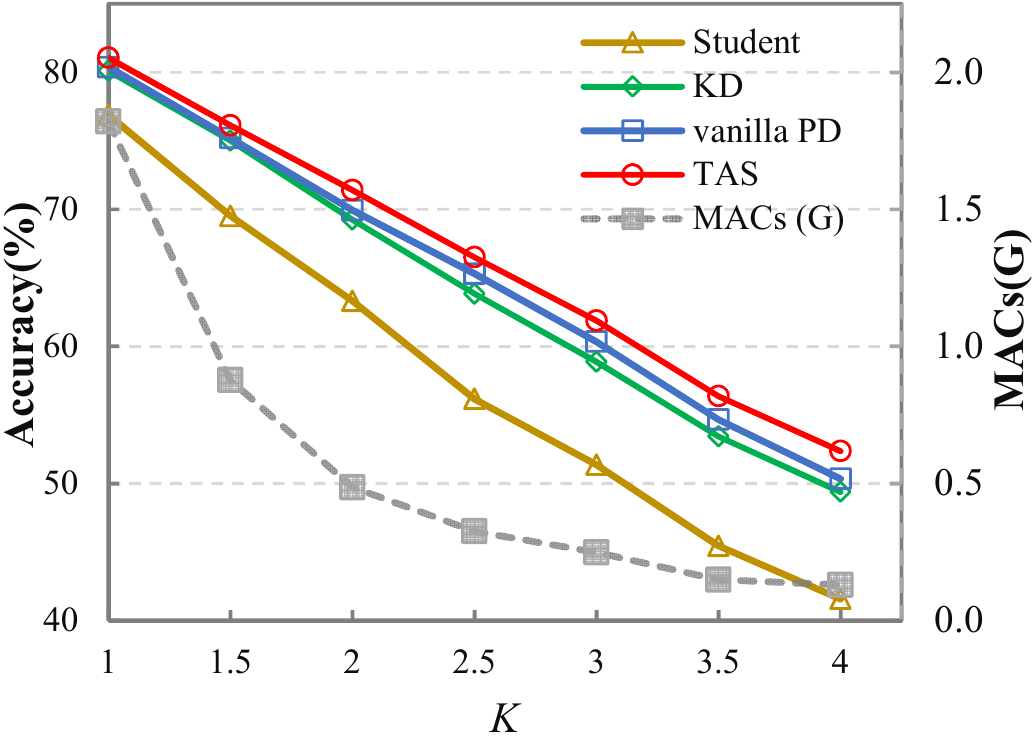}
	\end{center}
	\caption{Accuracy (\%) and MACs(G) of more input sizes on the setting (b) of CUB datasets.}
	\label{fig_k}
\end{figure}

\noindent \textbf{Teacher-student pairs. 
In Table~\ref{tab:setting}, we provide the detailed setting of our image classification task, which contains six teacher-student pairs with two down-sampling rate, including the model size (Params), computational complexity (MACs), and efficient ratio (Compute Red.). 
We use variants of ResNet~\cite{he2016deep}, ShuffleNetV2~\cite{ma2018shufflenet} and ViT~\cite{dosovitskiy2020vit} as the teacher and student models. In settings (a) to (c), both teacher and student belong to CNN. For settings (d) and (e), one of the teachers or students is CNN and another is ViT. For setting (f), both teacher and student models belong to ViT.
With regard to the input, the spatial size of the large input is $224\times 224$, while the spatial size of the small input is $\frac{224}{K}\times \frac{224}{K}$, we report the details of two down-sampling rates in Table~\ref{tab:setting}, \ie $K=2$ and $K=4$. Moreover, we report the performance of more down-sampling rates in Fig.~\ref{fig_k}.}

\begin{table}[!t]
	\centering
	\caption{Results on the CUB dataset for setting (d) to (e). The best is shown in bold. Each experiment is repeated five times and we report the mean value.}
	\resizebox{\linewidth}{!}{
		\begin{tabular}{l|cc|cc|cc}
			\hline
			& \multicolumn{2}{c|}{(d)} & \multicolumn{2}{c|}{(e)} & \multicolumn{2}{c}{(f)} \\
			& $K=2$  & $K=4$  & $K=2$  & $K=4$  & $K=2$  & $K=4$\\
			\hline
			Teacher & \multicolumn{2}{c|}{80.46 } & \multicolumn{2}{c|}{88.13 } & \multicolumn{2}{c}{88.13 } \\
			\hline
			Baseline-FS & 9.49  & 6.55  & 37.16  & 24.52  & 9.49  & 6.55  \\
			Baseline-FT & 69.14  & 32.14  & 63.31  & 41.56  & 69.14  & 32.14  \\
			\hline
			KD    & 71.67  & 39.05  & 71.31  & 51.28  & 74.26  & 40.02  \\
			DKD   & 70.94  & 38.29  & 70.66  & 49.09  & 73.41  & 40.92  \\
			\hline
			vanilla PD (One-Stage) & 72.41  & 40.06  & 71.90  & 51.77  & 74.67  & 40.58  \\
			TAS (Two-Stage) & \textbf{73.46} & \textbf{41.93} & \textbf{73.38} & \textbf{54.17} & \textbf{76.74} & \textbf{42.46} \\
			\hline
		\end{tabular}%
	}
	\label{tab:cub2}%
\end{table}%

\noindent\textbf{Experiments on CUB and Aircraft.} From Table~\ref{tab:cub} to~\ref{tab:aircraft2}, we report the performance for all settings on two fine-grained datasets. All the experiments are repeated five times and we report the mean value to avoid the influence of randomness on experimental results. We compared our methods with all aforementioned knowledge distillation methods for setting (a) to (c) where both teacher and student belong to CNN, and only prediction-based distillation methods are compared for setting (d) to (f) as ViT is involved. ``Baseline-FS'' denotes the student trained from scratch, and ``Baseline-FT'' denotes the student trained from pre-trained weights on ImageNet. Our proposed methods use KD as the prediction distillation method. From the results, we have the following observations:
\begin{enumerate}[label=\arabic*)]
	\item Our one-stage trained baseline vanilla PD can stably provide performance gains over KD on all settings, whether the student model belongs to CNN or ViT. Moreover, in most settings, our one-stage trained vanilla PD outperforms the knowledge distillation methods, which demonstrates the effectiveness of our proposed vanilla PD.
	\item The two-stage trained TAS framework enhances the performance of the baseline vanilla PD across all settings, providing additional performance gains for the student. This is because TAS reduces the learning difficulty of the student and introduces the feature distillation mechanism to relieve the performance degradation caused by the small input size. 
	\item Prediction-based distillation can provide stable performance gain for the student on all network architectures and input size settings. 
	\item The performance of feature distillation is very unstable: SP+KD and IC+KD can only provide a light performance gain over KD on some settings when $K$=2, and all three feature distillation methods perform badly when the input size is too low (\ie, $K$=4). This is because the teacher and student in pixel distillation have a larger gap than in knowledge distillation, which will make it difficult for the student to successfully mimic the teacher~\cite{cho2019efficacy}. 
	\item The performance of different models varies greatly with the change in input size and dataset. For instance, when $K$=4, student ViT-Ti/16 trained by ViT-B/16 (setting (f)) obtains the best performance on CUB (54.17\%), but student ResNet18 trained by ResNet50 is the best on the Aircraft dataset.
\end{enumerate}

\begin{table}[!t]
	\centering
	\footnotesize
	\caption{Results on the Aircraft dataset for setting (a) to (c). The best is shown in bold. Each experiment is repeated five times and we report the mean value.}
	\resizebox{\linewidth}{!}{
		\begin{tabular}{l|cc|cc|cc}
			\hline
			& \multicolumn{2}{c|}{(a)} & \multicolumn{2}{c|}{(b)} & \multicolumn{2}{c}{(c)} \\
			& $K=2$  & $K=4$  & $K=2$  & $K=4$  & $K=2$  & $K=4$\\
			\hline
			Teacher & \multicolumn{2}{c|}{86.26 } & \multicolumn{2}{c|}{85.30 } & \multicolumn{2}{c}{79.96 } \\
			\hline
			Baseline-FS & 51.52  & 32.37  & 50.31  & 33.52  & 45.78  & 26.73  \\
			Baseline-FT & 70.79  & 47.11  & 68.38  & 49.12  & 63.69  & 45.21  \\
			\hline
			KD    & 73.65  & 53.60  & 73.70  & 55.14  & 66.30  & 46.92  \\
			AT    & 70.08  & 7.73  & 68.36  & 9.29  & 43.43  & 8.08  \\
			AT+KD & 74.01  & 17.77  & 72.95  & 18.21  & 58.19  & 15.81  \\
			SP    & 71.69  & 41.40  & 70.15  & 39.68  & 61.19  & 10.76  \\
			SP+KD & 74.43  & 52.05  & 73.49  & 50.20  & 66.40  & 37.79  \\
			IC    & 71.81  & 46.35  & 70.97  & 47.01  & 62.43  & 42.61  \\
			IC+KD & 73.92  & 53.12  & 73.05  & 52.85  & 66.51  & 44.84  \\
			DKD   & 73.75  & 52.52  & 72.31  & 53.62  & 65.17  & 45.80  \\
			\hline
			vanilla PD (One-Stage) & 74.79  & 54.51  & 74.09  & 56.81  & 66.91  & 47.52  \\
			TAS (Two-Stage) & \textbf{76.51} & \textbf{56.17} & \textbf{75.72} & \textbf{58.31} & \textbf{67.84} & \textbf{48.67} \\
			\hline
		\end{tabular}%
	}
	\label{tab:aircraft}%
\end{table}%
\begin{table}[!t]
	\centering
	\caption{Results on the Aircraft dataset for setting (d) to (e). The best is shown in bold. Each experiment is repeated five times and we report the mean value.}
	\resizebox{\linewidth}{!}{
		\begin{tabular}{l|cc|cc|cc}
			\hline
			& \multicolumn{2}{c|}{(d)} & \multicolumn{2}{c|}{(e)} & \multicolumn{2}{c}{(f)} \\
			& $K=2$  & $K=4$  & $K=2$  & $K=4$  & $K=2$  & $K=4$\\
			\hline
			Teacher & \multicolumn{2}{c|}{85.30 } & \multicolumn{2}{c|}{79.60 } & \multicolumn{2}{c}{79.60 } \\
			\hline
			Baseline-FS & 9.38  & 5.71  & 50.31  & 33.52  & 9.38  & 5.71  \\
			Baseline-FT & 57.35  & 31.89  & 68.38  & 49.12  & 57.35  & 31.89  \\
			\hline
			KD    & 65.45  & 40.02  & 70.60  & 55.46  & 66.44  & 42.86  \\
			DKD   & 64.31  & 40.21  & 69.68  & 54.91  & 66.85  & 42.12  \\
			\hline
			vanilla PD (One-Stage) & 66.41  & 40.89  & 71.66  & 55.61  & 67.68  & 43.22  \\
			TAS (Two-Stage) & \textbf{67.78} & \textbf{41.88} & \textbf{72.74} & \textbf{57.54} & \textbf{68.85} & \textbf{44.65} \\
			\hline
		\end{tabular}%
	}
	\label{tab:aircraft2}%
\end{table}%

\noindent\textbf{Experiments on More Input Sizes.} As shown in Fig.~\ref{fig_k}, we conduct experiments on more small input sizes to demonstrate the generalization ability of our methods on the input size. The teacher is ResNet50 with $224\times 224$ input and the student is ResNet18. $K$ is set from 1 to 4 with stride 0.5. We can observe that both the one-stage vanilla PD and two-stage TAS can obtain performance gains on all sizes.

\noindent\textbf{Experiments on Imagenet.} In Table~\ref{tab:imagenet} we report the performance on the ImageNet dataset. All models are trained from scratch. We compared our methods with previous knowledge distillation methods for setting (b), \ie, the teacher is ResNet50 with input size $224\times 224$ and the student is ResNet18 with input size $112\times 112$ ($K$=2) and $56\times 56$ ($K$=4). We report our performance based on two prediction distillation methods KD and DKD. For KD based method, the one-stage trained vanilla PD can bring 0.52\% and 0.46\% performance gains when $K$ is 2 and 4, respectively. Also, when the prediction distillation method is DKD, vanilla PD can bring 0.34\% and 0.56\% performance gains when $K$ is 2 and 4, respectively. Using two-stage trained TAS with a prediction distillation method DKD achieves the best performance, which outperforms the baseline student by 2.32\% and 3.02\%.

\begin{table}[!t]
	\centering
	\caption{Results on and ImageNet dataset for setting (b). Teacher is \textbf{ResNet50} and student is \textbf{ResNet18}.}
\setlength{\tabcolsep}{4mm}{
      \renewcommand{\arraystretch}{1}{
		\begin{tabular}{l|c|c}
			\hline
			\multicolumn{1}{l}{Teacher} & \multicolumn{2}{c}{80.37 } \\
			\hline
			& $K=2$  & $K=4$\\
			Student & 62.04  & 50.81  \\
			\hline
			KD    & 62.70  & 51.41  \\
			AT    & 57.77  & 30.34  \\
			AT+KD & 58.18  & 35.34  \\
			SP    & 62.17  & 50.87  \\
			SP+KD & 62.34  & 51.49  \\
			IC    & 62.28  & 51.16  \\
			IC+KD & 62.59  & 51.70  \\
			DKD   & 63.56  & 52.28  \\
			\hline
			vanilla PD (KD) & 63.22  & 51.87  \\
			vanilla PD (DKD) & 63.90  & 52.84  \\
			TAS (KD) & 63.71  & 52.44  \\
			TAS (DKD) & \textbf{64.36} & \textbf{53.83} \\
			\hline
		\end{tabular}%
	}}
	\label{tab:imagenet}%
\end{table}%

\noindent \textbf{Comparision with super-resolution method.} At first glance, an intuitive solution to the challenge of recognizing low-resolution images involves upscaling the input to a higher resolution, and subsequently processing these upscaled inputs with a model that has been trained on large inputs to make predictions. With this perspective, we introduce two baseline models in Table~\ref{tab:sr} denoted as experiment (ii). To be specific, in experiments \textbf{i} of Table~\ref{tab:sr}, we provide the performance of the teacher network and the student network trained by HR images (Baseline-HR). Then, experiment (ii) represents the upscaling paradigms: pre-trained Baseline-HR models are used in inference, the LR inputs of students are upsampled into HR images via bilinear interpolation (Baseline-Bilinear) and the super-resolution model SwinIR-M~\cite{liang2021swinir}, respectively. In experiments (iii), we provide the performance of students trained by LR images (Baseline-LR) and our methods. We can observe that the upscaling paradigm in experiments (ii) can outperform models with LR input in most cases, but it has two drawbacks that make it fundamentally different from pixel distillation: \textbf{1)} it needs more computational complexity that is caused by the HR input of the student and the SR operation. Take setting (b) with $K=2$ as an example, when the input resolution of ResNet-18 increases from $112\times 112$ into $224\times 224$, the MACs will increase from 0.487G into 1.820G, and the SR model SwinIR-M causes another 8.38G costs. \textbf{2)} The process of generation and prediction of SR images cannot be calculated in parallel, which means it will inevitably lead to additional calculation time.

\begin{table}[!t]
  \centering
  \caption{Comparision with super-resolution method on CUB dataset with settings (b) and (f).}
    \begin{tabular}{c|c|cc|cc}
    \hline
    \textcolor[rgb]{ 0,  0,  0}{} & \textcolor[rgb]{ 0,  0,  0}{} & \multicolumn{2}{c|}{\textcolor[rgb]{ 0,  0,  0}{(b)}} & \multicolumn{2}{c}{\textcolor[rgb]{ 0,  0,  0}{(f)}} \bigstrut[t]\\
    \textcolor[rgb]{ 0,  0,  0}{\textbf{id}} & \textcolor[rgb]{ 0,  0,  0}{Method} & \textcolor[rgb]{ 0,  0,  0}{K=2} & \textcolor[rgb]{ 0,  0,  0}{K=4} & \textcolor[rgb]{ 0,  0,  0}{K=2} & \textcolor[rgb]{ 0,  0,  0}{K=4} \bigstrut[b]\\
    \hline
    \multirow{2}[2]{*}{\textcolor[rgb]{ 0,  0,  0}{\textbf{i}}} & \textcolor[rgb]{ 0,  0,  0}{Teacher} & \multicolumn{2}{c|}{\textcolor[rgb]{ 0,  0,  0}{80.46}} & \multicolumn{2}{c}{\textcolor[rgb]{ 0,  0,  0}{88.13}} \bigstrut[t]\\
          & \textcolor[rgb]{ 0,  0,  0}{Baseline-HR} & \multicolumn{2}{c|}{\textcolor[rgb]{ 0,  0,  0}{76.89}} & \multicolumn{2}{c}{\textcolor[rgb]{ 0,  0,  0}{83.19}} \bigstrut[b]\\
    \hline
    \multirow{2}[2]{*}{\textcolor[rgb]{ 0,  0,  0}{\textbf{ii}}} & \textcolor[rgb]{ 0,  0,  0}{Baseline-Bilinear} & \textcolor[rgb]{ 0,  0,  0}{70.92 } & \textcolor[rgb]{ 0,  0,  0}{62.45 } & \textcolor[rgb]{ 0,  0,  0}{80.87 } & \textcolor[rgb]{ 0,  0,  0}{75.91 } \bigstrut[t]\\
          & \textcolor[rgb]{ 0,  0,  0}{Baseline-SwinIR} & \textcolor[rgb]{ 0,  0,  0}{74.73 } & \textcolor[rgb]{ 0,  0,  0}{67.14 } & \textcolor[rgb]{ 0,  0,  0}{82.06 } & \textcolor[rgb]{ 0,  0,  0}{78.75 } \bigstrut[b]\\
    \hline
    \multirow{3}[2]{*}{\textcolor[rgb]{ 0,  0,  0}{\textbf{iii}}} & \textcolor[rgb]{ 0,  0,  0}{Baseline-LR} & \textcolor[rgb]{ 0,  0,  0}{63.31 } & \textcolor[rgb]{ 0,  0,  0}{41.56 } & \textcolor[rgb]{ 0,  0,  0}{69.14 } & \textcolor[rgb]{ 0,  0,  0}{32.14 } \bigstrut[t]\\
          & \textcolor[rgb]{ 0,  0,  0}{vanilla PD (One-Stage)} & \textcolor[rgb]{ 0,  0,  0}{69.94 } & \textcolor[rgb]{ 0,  0,  0}{50.34 } & \textcolor[rgb]{ 0,  0,  0}{74.67 } & \textcolor[rgb]{ 0,  0,  0}{40.58 } \\
          & \textcolor[rgb]{ 0,  0,  0}{TAS (Two-Stage)} & \textcolor[rgb]{ 0,  0,  0}{71.65 } & \textcolor[rgb]{ 0,  0,  0}{52.44 } & \textcolor[rgb]{ 0,  0,  0}{76.74 } & \textcolor[rgb]{ 0,  0,  0}{42.46 } \bigstrut[b]\\
    \hline
    \end{tabular}%
  \label{tab:sr}%
\end{table}%

\subsection{Experimental Results of Object Detection}

In this section, we compared our method with previous knowledge distillation methods in object detection, including prediction distillation method CrossKD~\cite{wang2023crosskd}, and the feature distillation methods proposed by Cao \etal ~\cite{cao2022pkd} that distillation from the FPN.

\begin{table}[!t]
  \centering
  \caption{The configuration of teacher and student models for four experiment settings of object detection. We report model size (Params), computational complexity (FLOPs), and computational complexity reduction (Compute Red.)}
  \resizebox{\linewidth}{!}{
    \begin{tabular}{c|lcccc}
    \hline
    \textcolor[rgb]{ 0,  0,  0}{} & \multicolumn{1}{l}{\textcolor[rgb]{ 0,  0,  0}{}} & \multicolumn{1}{c}{\textcolor[rgb]{ 0,  0,  0}{\textbf{(g)}}} & \multicolumn{1}{c}{\textcolor[rgb]{ 0,  0,  0}{\textbf{(h)}}} & \multicolumn{1}{c}{\textcolor[rgb]{ 0,  0,  0}{\textbf{(i)}}} & \multicolumn{1}{c}{\textcolor[rgb]{ 0,  0,  0}{\textbf{(j)}}} \bigstrut\\
    \hline
    \multicolumn{1}{c|}{\multirow{3}[4]{*}{\textcolor[rgb]{ 0,  0,  0}{Teacher}}} & \textcolor[rgb]{ 0,  0,  0}{Backbone} & \multicolumn{1}{c}{\textcolor[rgb]{ 0,  0,  0}{ResNet50}} & \multicolumn{1}{c}{\textcolor[rgb]{ 0,  0,  0}{Swin-T}} & \multicolumn{1}{c}{\textcolor[rgb]{ 0,  0,  0}{ResNet101}} & \multicolumn{1}{c}{\textcolor[rgb]{ 0,  0,  0}{Swin-B}} \bigstrut[t]\\
          & \textcolor[rgb]{ 0,  0,  0}{Param. (M)} & \textcolor[rgb]{ 0,  0,  0}{36.7 } & \textcolor[rgb]{ 0,  0,  0}{38.5 } & \textcolor[rgb]{ 0,  0,  0}{55.7 } & \textcolor[rgb]{ 0,  0,  0}{98.4 } \bigstrut[b]\\
\cline{2-6}          & \textcolor[rgb]{ 0,  0,  0}{FLOPs (G)} & \textcolor[rgb]{ 0,  0,  0}{107 } & \textcolor[rgb]{ 0,  0,  0}{125 } & \textcolor[rgb]{ 0,  0,  0}{145 } & \textcolor[rgb]{ 0,  0,  0}{245 } \bigstrut\\
    \hline
    \multicolumn{1}{c|}{\multirow{2}[2]{*}{\textcolor[rgb]{ 0,  0,  0}{Student}}} & \textcolor[rgb]{ 0,  0,  0}{Backbone} & \multicolumn{1}{c}{\textcolor[rgb]{ 0,  0,  0}{ResNet18}} & \multicolumn{1}{c}{\textcolor[rgb]{ 0,  0,  0}{ResNet18}} & \multicolumn{1}{c}{\textcolor[rgb]{ 0,  0,  0}{Swin-T}} & \multicolumn{1}{c}{\textcolor[rgb]{ 0,  0,  0}{Swin-T}} \bigstrut[t]\\
          & \textcolor[rgb]{ 0,  0,  0}{Param. (M)} & \textcolor[rgb]{ 0,  0,  0}{22.1 } & \textcolor[rgb]{ 0,  0,  0}{22.1 } & \textcolor[rgb]{ 0,  0,  0}{38.5 } & \textcolor[rgb]{ 0,  0,  0}{38.5 } \bigstrut[b]\\
    \hline
    \multicolumn{1}{c|}{\multirow{2}[2]{*}{\textcolor[rgb]{ 0,  0,  0}{$K$=2}}} & \textcolor[rgb]{ 0,  0,  0}{FLOPs (G)} & \textcolor[rgb]{ 0,  0,  0}{20.2 } & \textcolor[rgb]{ 0,  0,  0}{20.2 } & \textcolor[rgb]{ 0,  0,  0}{34.2 } & \textcolor[rgb]{ 0,  0,  0}{34.2 } \bigstrut[t]\\
          & \textcolor[rgb]{ 0,  0,  0}{Compute Red.} & \textcolor[rgb]{ 0,  0,  0}{81.2\%} & \textcolor[rgb]{ 0,  0,  0}{83.9\%} & \textcolor[rgb]{ 0,  0,  0}{76.4\%} & \textcolor[rgb]{ 0,  0,  0}{86.0\%} \bigstrut[b]\\
    \hline
    \multicolumn{1}{c|}{\multirow{2}[2]{*}{\textcolor[rgb]{ 0,  0,  0}{$K$=4}}} & \textcolor[rgb]{ 0,  0,  0}{FLOPs (G)} & \textcolor[rgb]{ 0,  0,  0}{5.9 } & \textcolor[rgb]{ 0,  0,  0}{5.9 } & \textcolor[rgb]{ 0,  0,  0}{9.4 } & \textcolor[rgb]{ 0,  0,  0}{9.4 } \bigstrut[t]\\
          & \textcolor[rgb]{ 0,  0,  0}{Compute Red.} & \textcolor[rgb]{ 0,  0,  0}{94.5\%} & \textcolor[rgb]{ 0,  0,  0}{95.3\%} & \textcolor[rgb]{ 0,  0,  0}{93.5\%} & \textcolor[rgb]{ 0,  0,  0}{96.2\%} \bigstrut[b]\\
    \hline
    \end{tabular}%
    }
  \label{tab:settingod}%
\end{table}%

\noindent \textbf{Teacher-student pairs.} 
In Table~\ref{tab:settingod}, we provide the detailed setting of the object detection task, which contains four teacher-student pairs with two down-sampling rates, including the model size (Params), computational complexity (FLOPs), and efficient ratio. Floating point operations (FLOPs) are used for computing computational complexity because we use the officially provided tools of MMDetection\footnote{\url{https://github.com/open-mmlab/mmdetection/blob/main/tools/analysis_tools/get_flops.py}}.
We use variants of ResNet~\cite{he2016deep}, Swin transformer~\cite{liu2021swin} as the teacher and student models. In settings (g), both teacher and student belong to ResNet. For settings (h) and (i), one of the teachers or students is ResNet and another is Swin transformer. For setting (j), both teacher and student models belong to Swin transformer.
In Table~\ref{tab:settingod}, we report all details on the PASCAL VOC dataset, where the spatial size of the large input is $1000\times 600$, and that of the small input is $\frac{1000}{K}\times \frac{600}{K}$, we report the details of two down-sampling rates, \ie $K=2$ and $K=4$. For the COCO dataset, the spatial size of the large input is $1333\times 800$. All the models utilize the RetinaNet~\cite{lin2017focal} framework. 

\begin{table}[!t]
  \centering
  \caption{Results (mAP) on the PASCAL VOC dataset from setting (g) to (j). The best is shown in bold.}
    \begin{tabular}{l|cc|cc}
    \hline
    \textcolor[rgb]{ 0,  0,  0}{\textit{\textbf{CNN-based Student}}} & \multicolumn{2}{c|}{\textcolor[rgb]{ 0,  0,  0}{\textbf{(g)}}} & \multicolumn{2}{c}{\textcolor[rgb]{ 0,  0,  0}{\textbf{(h)}}} \bigstrut\\
    \hline
    \textcolor[rgb]{ 0,  0,  0}{Teacher} & \multicolumn{2}{c|}{\textcolor[rgb]{ 0,  0,  0}{76.9}} & \multicolumn{2}{c}{\textcolor[rgb]{ 0,  0,  0}{79.6}} \bigstrut\\
    \hline
    \textcolor[rgb]{ 0,  0,  0}{} & \textcolor[rgb]{ 0,  0,  0}{$K$=2} & \textcolor[rgb]{ 0,  0,  0}{$K$=4} & \textcolor[rgb]{ 0,  0,  0}{$K$=2} & \textcolor[rgb]{ 0,  0,  0}{$K$=4} \bigstrut[t]\\
    \textcolor[rgb]{ 0,  0,  0}{Student} & \textcolor[rgb]{ 0,  0,  0}{69.0 } & \textcolor[rgb]{ 0,  0,  0}{56.9 } & \textcolor[rgb]{ 0,  0,  0}{69.0 } & \textcolor[rgb]{ 0,  0,  0}{56.9 } \bigstrut[b]\\
    \hline
    \textcolor[rgb]{ 0,  0,  0}{TAS-AFP (CrossKD)} & \textcolor[rgb]{ 0,  0,  0}{72.3 } & \textcolor[rgb]{ 0,  0,  0}{60.3 } & \textcolor[rgb]{ 0,  0,  0}{72.6 } & \textcolor[rgb]{ 0,  0,  0}{61.3 } \bigstrut[t]\\
    \textcolor[rgb]{ 0,  0,  0}{TAS-AFP (CrossKD) + Cao \etal} & \textcolor[rgb]{ 0,  0,  0}{72.5 } & \textcolor[rgb]{ 0,  0,  0}{60.3 } & \textcolor[rgb]{ 0,  0,  0}{73.4 } & \textcolor[rgb]{ 0,  0,  0}{61.1 } \\
    \textcolor[rgb]{ 0,  0,  0}{TAS-AFP (CrossKD) + ISRD} & \textcolor[rgb]{ 0,  0,  0}{\textbf{72.9 }} & \textcolor[rgb]{ 0,  0,  0}{\textbf{61.1 }} & \textcolor[rgb]{ 0,  0,  0}{\textbf{73.9 }} & \textcolor[rgb]{ 0,  0,  0}{\textbf{62.9 }} \bigstrut[b]\\
    \hline
    \hline
    \textcolor[rgb]{ 0,  0,  0}{\textit{\textbf{Swin-based Student}}} & \multicolumn{2}{c|}{\textcolor[rgb]{ 0,  0,  0}{\textbf{(i)}}} & \multicolumn{2}{c}{\textcolor[rgb]{ 0,  0,  0}{\textbf{(j)}}} \bigstrut\\
    \hline
    \textcolor[rgb]{ 0,  0,  0}{Teacher} & \multicolumn{2}{c|}{\textcolor[rgb]{ 0,  0,  0}{78.4}} & \multicolumn{2}{c}{\textcolor[rgb]{ 0,  0,  0}{83.4}} \bigstrut\\
    \hline
    \textcolor[rgb]{ 0,  0,  0}{} & \textcolor[rgb]{ 0,  0,  0}{$K$=2} & \textcolor[rgb]{ 0,  0,  0}{$K$=4} & \textcolor[rgb]{ 0,  0,  0}{$K$=2} & \textcolor[rgb]{ 0,  0,  0}{$K$=4} \bigstrut[t]\\
    \textcolor[rgb]{ 0,  0,  0}{Student} & \textcolor[rgb]{ 0,  0,  0}{77.4 } & \textcolor[rgb]{ 0,  0,  0}{66.8 } & \textcolor[rgb]{ 0,  0,  0}{77.4 } & \textcolor[rgb]{ 0,  0,  0}{66.8 } \bigstrut[b]\\
    \hline
    \textcolor[rgb]{ 0,  0,  0}{TAS-AFP (CrossKD)} & \textcolor[rgb]{ 0,  0,  0}{77.2 } & \textcolor[rgb]{ 0,  0,  0}{67.3 } & \textcolor[rgb]{ 0,  0,  0}{78.3 } & \textcolor[rgb]{ 0,  0,  0}{69.4 } \bigstrut[t]\\
    \textcolor[rgb]{ 0,  0,  0}{TAS-AFP (CrossKD) + Cao \etal} & \textcolor[rgb]{ 0,  0,  0}{76.7 } & \textcolor[rgb]{ 0,  0,  0}{67.6 } & \textcolor[rgb]{ 0,  0,  0}{78.6 } & \textcolor[rgb]{ 0,  0,  0}{68.9 } \\
    \textcolor[rgb]{ 0,  0,  0}{TAS-AFP (CrossKD) + ISRD} & \textcolor[rgb]{ 0,  0,  0}{\textbf{78.0 }} & \textcolor[rgb]{ 0,  0,  0}{\textbf{68.7 }} & \textcolor[rgb]{ 0,  0,  0}{\textbf{78.9 }} & \textcolor[rgb]{ 0,  0,  0}{\textbf{70.5 }} \bigstrut[b]\\
    \hline
    \end{tabular}%
  \label{tab:res_voc}%
\end{table}%

\begin{table}[!t]
  \centering
  \caption{Results (mAP) on the COCO dataset for setting (h) and (i). The best is shown in bold.}
    \begin{tabular}{l|cc|cc}
    \hline
    \textcolor[rgb]{ 0,  0,  0}{} & \multicolumn{2}{c|}{\textcolor[rgb]{ 0,  0,  0}{(h)}} & \multicolumn{2}{c}{\textcolor[rgb]{ 0,  0,  0}{(i)}} \bigstrut\\
    \hline
    \textcolor[rgb]{ 0,  0,  0}{Teacher} & \multicolumn{2}{c|}{\textcolor[rgb]{ 0,  0,  0}{37.3}} & \multicolumn{2}{c}{\textcolor[rgb]{ 0,  0,  0}{38.7}} \bigstrut\\
    \hline
    \textcolor[rgb]{ 0,  0,  0}{} & \textcolor[rgb]{ 0,  0,  0}{$K$=2} & \textcolor[rgb]{ 0,  0,  0}{$K$=4} & \textcolor[rgb]{ 0,  0,  0}{$K$=2} & \textcolor[rgb]{ 0,  0,  0}{$K$=4} \bigstrut[t]\\
    \textcolor[rgb]{ 0,  0,  0}{Student} & \textcolor[rgb]{ 0,  0,  0}{27.2 } & \textcolor[rgb]{ 0,  0,  0}{18.7 } & \textcolor[rgb]{ 0,  0,  0}{32.5 } & \textcolor[rgb]{ 0,  0,  0}{23.5 } \bigstrut[b]\\
    \hline
    \textcolor[rgb]{ 0,  0,  0}{TAS-AFP (CrossKD)} & \textcolor[rgb]{ 0,  0,  0}{30.2 } & \textcolor[rgb]{ 0,  0,  0}{21.1 } & \textcolor[rgb]{ 0,  0,  0}{33.7 } & \textcolor[rgb]{ 0,  0,  0}{25.1 } \bigstrut[t]\\
    \textcolor[rgb]{ 0,  0,  0}{TAS-AFP (CrossKD) + Cao \etal} & \textcolor[rgb]{ 0,  0,  0}{30.4 } & \textcolor[rgb]{ 0,  0,  0}{20.9 } & \textcolor[rgb]{ 0,  0,  0}{35.0 } & \textcolor[rgb]{ 0,  0,  0}{24.6 } \\
    \textcolor[rgb]{ 0,  0,  0}{TAS-AFP (CrossKD) + ISRD} & \textcolor[rgb]{ 0,  0,  0}{\textbf{30.6 }} & \textcolor[rgb]{ 0,  0,  0}{\textbf{22.1 }} & \textcolor[rgb]{ 0,  0,  0}{\textbf{35.6 }} & \textcolor[rgb]{ 0,  0,  0}{\textbf{25.9 }} \bigstrut[b]\\
    \hline
    \end{tabular}%
  \label{tab:res_coco}%
\end{table}%

\noindent \textbf{Experiments on PASCAL VOC and COCO.} Table~\ref{tab:res_voc} and ~\ref{tab:res_coco} report performance on the PASCAL VOC and COCO dataset, respectively. \textbf{TAS-AFP (CrossKD)} is a base method that is built by our TAS framework and AFP strategy, and the prediction-based distillation CrossKD~\cite{wang2023crosskd} is utilized, we can observe that this base model can bring performance for all settings on both datasets. For example, on the setting (h) of PASCAL VOC, it can bring 4.4\% mAP improvement when $K$=4 (56.9 vs. 61.3). Then, \textbf{TAS-AFP (CrossKD)+Cao \etal} introduce the FPN based feature distillation method~\cite{cao2022pkd}, this method is not stable when the resolution is reduced. on the setting (i) of COCO with $K$=2, it can improve the mAP from 33.7\% to 35.0\%, but it brings performance degradation in most cases when $K$=4. Finally, \textbf{TAS-AFP (CrossKD)+ISRD} indicates we use our ISRD mechanism in the input compression stage, which can bring stable performance improvements under all settings.

\subsection{Ablation Studies and Model Analysis}
\label{sec:ablation}
In this section, we provide ablation studies about the key components, the channel expanding layer of ISRD, the architecture of TAS, and hyperparameters.

\noindent\textbf{Ablation study of the key learning components.} As shown in Table \ref{tab:ablation}, we conducted our ablation study on the CUB dataset. We select two settings, \ie, setting (b) whose student is CNN, and setting (f) whose student is ViT. The experimental results show that there will be a large performance gain via directly using traditional knowledge distillation (KD)~\cite{hinton2015distilling} approach, which can improve the accuracy from 41.56\% to 49.38\% on setting (b) when $K$=4. Then, ISRD can provide 0.56\% to 0.96\% performance gain for all settings, by integrating KD and our proposed ISRD we could obtain a one-stage trained simple baseline that can provide stable performance improvement for students. Furthermore, introducing the assistant network to achieve a two-stage trained framework can bring stable performance gains as it can reduce the learning difficulty of the student. Finally, the simple feature distillation in the input compression stage (\ie, ICF) can further reduce the performance degradation caused by the small input size. Such performance gain demonstrates that our proposed learning framework can work effectively with lightweight network architecture and small input sizes.

\begin{table}[!t]
	\centering
	\caption{Ablation study of each component on setting (b) and (f) in the image classification task, we report performance on the CUB dataset.}
	\resizebox{1\linewidth}{!}{
		\begin{tabular}{ccc|cc|cc|cc}
			\hline
			\multicolumn{3}{c|}{One-stage} & \multicolumn{2}{c|}{Two-Stage} & \multicolumn{2}{c|}{(b)} & \multicolumn{2}{c}{(f)} \\
			Baseline & KD    & ISFR  & TAS   & ICF   & $K=2$  & $K=4$  & $K=2$  & $K=4$\\
			\hline
			$\surd$ &       &       &       &       & 63.31  & 41.56  & 69.14  & 32.14  \\
			$\surd$ & $\surd$ &       &       &       & 69.26  & 49.38  & 74.26  & 40.02  \\
			$\surd$ & $\surd$ & $\surd$ &       &       & 69.94  & 50.34  & 74.67  & 40.58  \\
			$\surd$ & $\surd$ & $\surd$ & $\surd$ &       & 70.31  & 51.40  & 75.87  & 41.64  \\
			$\surd$ & $\surd$ & $\surd$ & $\surd$ & $\surd$ & 71.65  & 52.44  & 76.79  & 42.46  \\
			\hline
		\end{tabular}%
	}
	\label{tab:ablation}%
\end{table}%

\begin{figure}[!t]
	\begin{center}
		\includegraphics[width=0.95\linewidth]{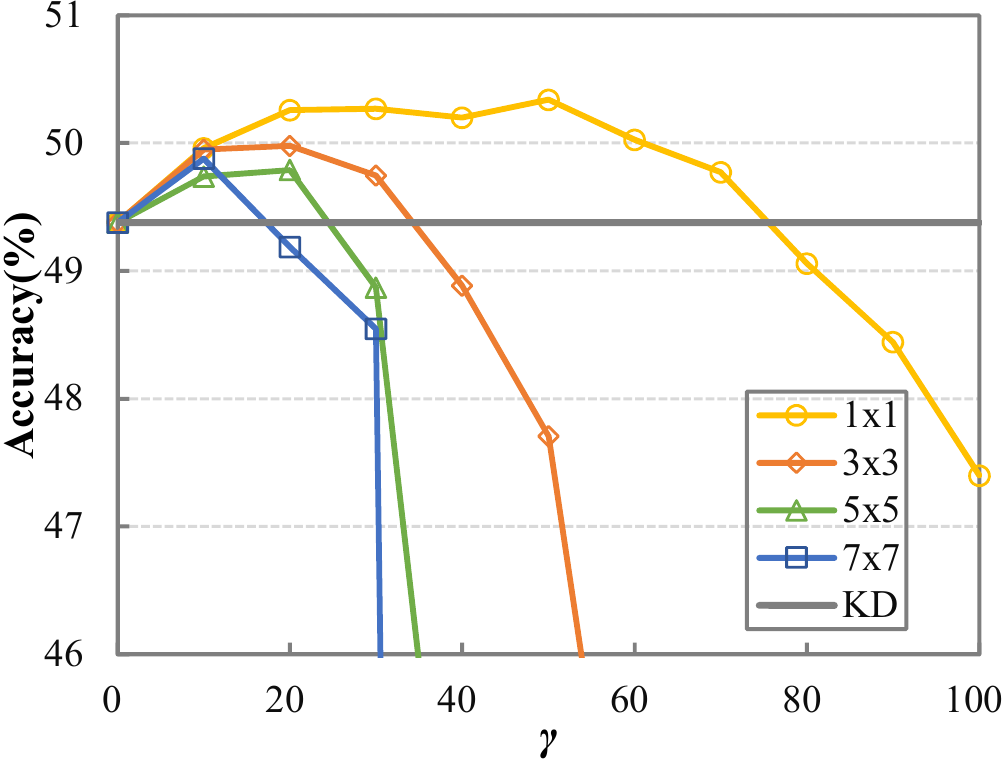}
	\end{center}
	\caption{Ablation study about the kernel size of the channel expanding layer in ISRD.}
	\label{fig_kernel}
\end{figure}

\noindent\textbf{Ablation study about the kernel size of the channel expanding layer in ISRD}. In the decoder of the proposed ISRD, a $1\times 1$ convolution layer is used and expand the volume of the input feature, a natural question about this process is whether using a larger kernel size can help obtain higher performance. As shown in Fig.~\ref{fig_kernel}, on setting (b) of the CUB dataset, we conduct experiments about kernel size from $1\times 1$ to $7\times 7$. To avoid the influence of hyperparameter $\gamma$, experiments of each kernel size are conducted when $\gamma$ increases from 10 to 100 with stride 10. We can observe that performance decreases when the kernel size increases from $1\times 1$ to $7\times 7$. The main reason is that the purpose of the ISRD is to transfer knowledge to the input module of the student, \ie, the only one learnable parameter in the encoder of the ISRD, even though using stronger decoder can help the ISRD to obtain better pseudo large images, but the information received by encoders will become weaker and further lead to lower performance gains for the student.

\noindent\textbf{Ablation studies about the architecture of the TAS}. TAS separates the pixel distillation into model compression and input compression, these two compression process can exchange their order. In Table~\ref{tab:tas}, we conduct experiments on settings (b) and (f) of CUB to explore the influence of this order. (1) is the traditional teacher-student framework, (2) and (3) are the proposed TAS framework. In experiment (2) we first perform the input compression process (vanilla PD) and then use the prediction distillation method KD to perform the model compression process. In experiment (3) we exchange the order of input compression and model compression. We can observe that the performance is better when we perform the model compression first, this is because input resolution has a greater impact on the performance. Hence, the performance of the assistant is too low if input compression is performed first.

\begin{table}[!t]
	\centering
	\caption{Ablation study about the TAS framework in image classification. We provide results on the CUB dataset with settings (b) and (f).}
	\resizebox{1\linewidth}{!}{
		\begin{tabular}{cc|cc|cc}
			\hline
			& Method & \multicolumn{2}{c|}{(b)} & \multicolumn{2}{c}{(f)} \\
			&       & $K=2$  & $K=4$  & $K=2$  & $K=4$\\
			\hline
			(1)   & vanilla PD (TS) & 69.94 & 50.34 & 74.67 & 40.58 \\
			(2)   & vanilla PD$\rightarrow$KD (TAS) & 69.74  & 49.74  & 75.21 & 40.861 \\
			(3)   & KD$\rightarrow$vanilla PD (TAS) & 70.31  & 51.40  & 75.87 & 41.636 \\
			\hline
		\end{tabular}%
	}
	\label{tab:tas}%
\end{table}%
\begin{figure*}[!t]
	\centering
	\begin{subfigure}{0.31\linewidth}
		\includegraphics[width=1\linewidth]{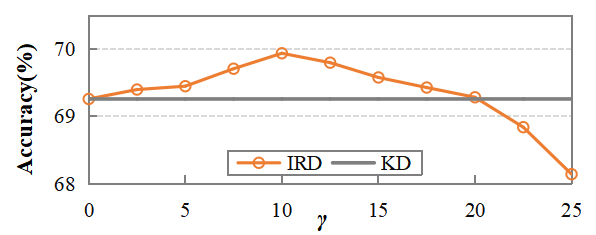}
		\caption{$K$=2, setting (b), ResNet18}
		\label{fig:k2b}
	\end{subfigure}
	\hfill
	\begin{subfigure}{0.31\linewidth}
		\includegraphics[width=1\linewidth]{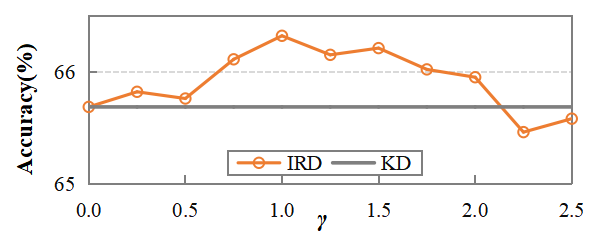}
		\caption{$K$=2, setting (c), ShuffleNetV2 1.0}
		\label{fig:k2c}
	\end{subfigure}
	\hfill
	\begin{subfigure}{0.31\linewidth}
		\includegraphics[width=1\linewidth]{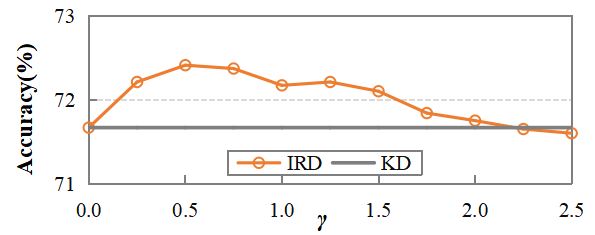}
		\caption{$K$=2, setting (d), ViT-Ti/16}
		\label{fig:k2d}
	\end{subfigure}
	\begin{subfigure}{0.31\linewidth}
		\includegraphics[width=1\linewidth]{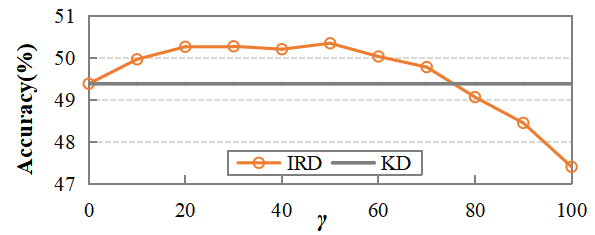}
		\caption{$K$=4, setting (b), student ResNet18}
		\label{fig:k4b}
	\end{subfigure}
	\hfill
	\begin{subfigure}{0.31\linewidth}
		\includegraphics[width=1\linewidth]{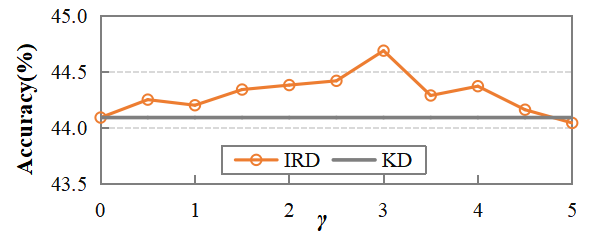}
		\caption{$K$=4, setting (c), ShuffleNetV2 1.0}
		\label{fig:k4c}
	\end{subfigure}
	\hfill
	\begin{subfigure}{0.31\linewidth}
		\includegraphics[width=1\linewidth]{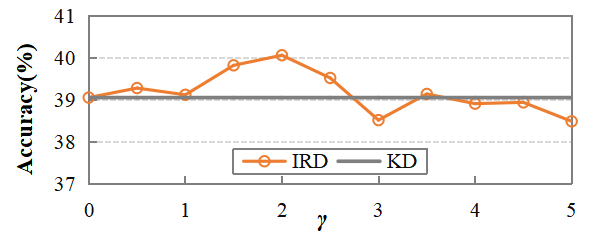}
		\caption{$K$=4, setting (d), ViT-Ti/16}
		\label{fig:k4d}
	\end{subfigure}
	\caption{Ablation study about the hyperparameter $\gamma$ on the CUB dataset under different teacher-student pairs and input sizes. For all experiments, the input size of the teacher and student is $224\times 224$ and $\frac{224}{K}\times \frac{224}{K}$, respectively, In the caption of each figure, we provide \textbf{the value of $K$, the index of setting, and the architecture of the student} from left to right.
 }
	\label{fig:gamma}
\end{figure*}

\noindent\textbf{Ablation studies about hyperparameter $\gamma$}. $\gamma$ is the only hypermeter for the one-stage trained vanilla PD. As shown in Fig.~\ref{fig:gamma}, we conduct experiments about $\gamma$ on setting (b) to (d) of the CUB dataset, their student is ResNet18, ShuffleNetV2 and ViT-Ti/16, respectively. We find the choice of $\gamma$ is high relative to two factors, the kernel size of the input convolution layer and the volume of the input feature: 
\begin{enumerate}[label=\arabic*)]
	\item The kernel size of the input convolution layer determines how many parameters can be used to learn knowledge from the large images. When the kernel size of the input feature is large, we can set a large $\gamma$ to obtain a better performance, and vice versa. For example, as illustrated in Fig.~\ref{fig:k2b} and Fig.~\ref{fig:k2c}, the kernel size of the input convolution layer of the ResNet18 is $3\times 7\times 7\times 64$, and the input convolution layer of the ShuffleNetV2 1.0 is $3\times 3\times 3\times 24$. When the input size for them is same ($112\times 112$), the best $\gamma$ for ResNet18 is 10 and the best $\gamma$ for ShuffleNetV2 1.0 is 1.0. 
	\item The volume of the input feature determines how well the quality of the pseudo large images. When the kernel size and the volume of the input feature are large, we can use a large $\gamma$ to obtain a better performance, and vice versa. One factor that affects the volume of the input feature is the network architecture: the input feature's volume of CNN is usually much larger than that of the ViT. For example, as shown in Fig.~\ref{fig:k2b} and Fig.~\ref{fig:k2d}, the input feature's volume of the ResNet18 and ViT-Ti/16 is $28\times 28\times 64$ and $49\times 192$, respectively. When both the input size ($112\times 112$) and teacher model (ResNet50) are the same for them, the best $\gamma$ for ResNet18 is 10 and the best $\gamma$ for ViT-Ti/16 is 0.5. Another factor that affects the volume of the input feature is the input size, we can observe that the best $\gamma$ in the first line of Fig.~\ref{fig:gamma} is smaller than that in the second line, this is because the input size of the first line is larger ($K=2$ \textit{vs.} $K=4$).
\end{enumerate}

\section{Conclusion}
In this paper, we propose a novel pixel distillation that aims to distill knowledge from a teacher model with heavy architecture and large input size to student models that have a variety of lightweight network architectures and input of different small sizes, which can provide more flexible cost control schemes than traditional knowledge distillation scheme.
We first provide a simple one-stag trained baseline for the classification task named vanilla PD, which can be adapted to sizes and different networks including CNN and ViT. Specifically, vanilla PD consists of a prediction-based distillation mechanism and a novel proposed input spatial representation distillation (ISRD) mechanism. ISRD can relieve the performance degradation due to the small input size by transferring information from the large inputs. Then we propose a teacher-assistant-student (TAS) framework to reduce the learning difficulty of students caused by the large gap between the teacher and student. TAS can also make it easier to relieve the performance degradation caused by small images by distilling knowledge in intermediate features. 
Experimental results demonstrate that the proposed method can improve the performance of models with various compact network architectures and small input sizes. Finally, we also apply the pixel distillation paradigm to a complex task, \ie, object detection, to showcase its potential for application in more scenarios.
In this phase, an Aligned Feature for Preservation (AFP) strategy is designed on the assistant network, which aligns the output dimensions of detectors at each stage by manipulating the scale of features before the detection head of the assistant network. In the future, we will apply the proposed distillation mechanism to other knowledge transfer tasks like~\cite{cheng2024continual,zhang2023weakly,li2023gp}

{\small
	\bibliographystyle{IEEEtran}
	\bibliography{egbib}
}

\end{document}